\documentclass{article}

    \PassOptionsToPackage{numbers, sort, compress}{natbib}

 \usepackage[eandd, final]{neurips_2026}

\usepackage[utf8]{inputenc} 
\usepackage[T1]{fontenc}    
\usepackage[
    colorlinks=true,
    linkcolor=blue!55!black,
    citecolor=green!45!black,
    urlcolor=magenta!55!black,
    filecolor=magenta!55!black,
    pdfborder={0 0 0}
]{hyperref}
\usepackage{url}            
\usepackage{booktabs}       
\usepackage{amsfonts}       
\usepackage{nicefrac}       
\usepackage{microtype}      
\usepackage{xcolor}         
\usepackage{pifont}
\usepackage{adjustbox}
\usepackage{graphicx}
\usepackage{subcaption}
\usepackage{amsmath}
\usepackage{tabularx}
\usepackage{multirow}
\usepackage[most]{tcolorbox}
\usepackage{hyperref}

\title{\texttt{Animation2Code}: Evaluating Temporal Visual Reasoning in Video-to-Code Generation}

\author{
  Anya Ji, Abhijith Varma Mudunuri, David M. Chan, Alane Suhr \\
  University of California, Berkeley\\
  \texttt{\{anyaji, mabhi02, davidchan, suhr\}@berkeley.edu} \\
}

\begin{document}
\maketitle

\begin{abstract}
  While recent vision-language models (VLMs) have achieved significant improvements on static visual-to-code tasks such as generating code for webpages, charts, or SVGs, it remains unclear whether they can recover temporal dynamics with code when motion is present. To address this gap, we introduce \texttt{Animation2Code}, a benchmark for evaluating temporal visual reasoning via reconstructing executable web animation code from videos. \texttt{Animation2Code} consists of 1,069 web animation videos with diverse visual appearances and motion patterns, paired with corresponding HTML/CSS/JavaScript implementations. We propose two human-aligned metrics, appearance similarity and temporal similarity, which allow us to disentangle visual fidelity from temporal alignment when comparing animations rendered from generated code against ground-truth renders. Evaluating state-of-the-art VLMs on this dataset shows that current VLMs struggle to maintain temporal consistency in reconstruction, even when achieving high appearance similarity, including under fine-tuning and iterative refinement settings. 
  Code and data are available at \href{https://anya-ji.github.io/animation2code-website/}{anya-ji.github.io/animation2code-website}.
\end{abstract}
\section{Introduction}
\begin{figure}[t]
\centering
\includegraphics[width=\linewidth]{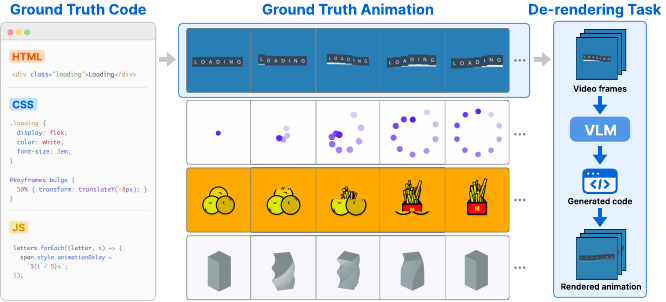}
\caption{\texttt{Animation2Code} is the \textbf{first benchmark to evaluate de-rendering of web animations to executable code}. \texttt{Animation2Code} comprises 1,069 high-quality web animation videos (four examples shown, five uniformly sampled frames each from videos rendered at 30 FPS) paired with HTML/CSS/JavaScript code (right). }
\label{fig:teaser}
\end{figure}

Visual perception involves not only what is happening at a particular moment in time, but also how sequences of moments connect to one another through temporal dynamics. Consider the case of web animations, where animations are used to guide user attention and present complex visuals: understanding a sequence of frames requires, for example, tracking objects that appear across multiple frames, their relative motion, and any emergent apparent motion, such as the ``wave'' effect in Figure~\ref{fig:teaser} (top example in the middle). 

While vision-language models (VLMs) have achieved strong performance on \textit{static} de-rendering tasks that map images of webpages, charts, formulas, or presentation slides to executable programs (e.g., \autoref{tab:dataset_comparison}; \cite{si2025design2code, wu2025plot2code, roberts2024image2struct,ge2025autopresent}), dynamic visual perception through de-rendering remains largely unexplored. This is fundamentally more challenging than static de-rendering, as it requires both reasoning over temporal dynamics, such as trajectory of objects and timing of motion, and translating these continuous dynamics into a discrete programmatic representation. 

To address this gap, we introduce \texttt{Animation2Code}, a benchmark that measures dynamic visual perception through success in a \textit{de-rendering} task.
During de-rendering, a reference video is mapped to an executable program that, when executed, results in a rendered video as similar as possible to the reference video in both appearance and dynamic motion.
Compared to existing benchmarks of video understanding which pose tasks like question-answering (e.g., \cite{fu2025video,li2024mvbench,li2024videovista, ning2025video}), successful de-rendering to executable code requires complete understanding of appearance and motion features, and results in an artifact which can be deterministically executed and rendered, and easily edited and adapted.

Besides a dataset of 1,069 web animation videos paired with ground-truth executable code, \texttt{Animation2Code} also contains a suite of metrics that support automatic evaluation of de-rendering. 
High-quality animation de-rendering requires correctness along multiple dimensions: the generated code must be executable, and when executed, the rendered video must resemble the target both in visual appearance and dynamic motion. 
We propose two automated metrics: appearance similarity (computed using frame alignment and image-level embedding similarity) and temporal similarity (computed using motion trajectory comparison).
We validate that both metrics correlate well with human judgments of both appearance and temporal similarity.

Through experiments across state-of-the-art commercial and open-source multimodal models (Gemini 3 Flash Preview, Qwen3-VL-8B-Instruct, GPT-5.4, Claude Sonnet 4.6, and LLaMA 4 Scout), including zero-shot prompting, supervised fine-tuning, and iterative refinement, we find a consistent gap: while models achieve strong appearance similarity, they struggle to reproduce precise temporal dynamics. This reveals a fundamental limitation in models' ability to interpret and replicate motion in visuals via code generation.

Our contributions are summarized as follows:
\begin{itemize}
    \item We introduce \texttt{Animation2Code} (\autoref{fig:teaser}), the first benchmark for dynamic visual machine perception through de-rendering web animations into executable code, consisting of 1,069 real-world animation video-code pairs with diverse motion patterns.
    \item We propose a human-aligned automatic evaluation suite that measures de-rendering quality by disentangling \textit{appearance} and \textit{temporal} similarity of rendered animations.
    \item We show that while state-of-the-art VLMs achieve near-perfect code execution (up to 100.0\%) and strong appearance reconstruction (up to 0.84 similarity), temporal similarity remains low across all models and settings (at most 0.32), revealing a key limitation in reasoning about temporal dynamics (\autoref{tab:results}).
\end{itemize}

\begin{table} 
\centering
\small
\caption{Comparison with representative visual-to-code benchmarks. 
\textbf{Our benchmark is the first to target de-rendering animation to code.} }
\label{tab:dataset_comparison}
\begin{tabularx}{\linewidth}{l c c c c c}
\toprule
\textbf{Benchmark} & \textbf{\#Samples} & \textbf{Input} & \textbf{Eval} & \textbf{Temporal Behavior} & \textbf{Target Domain} \\
\midrule
Design2Code \cite{si2025design2code}             
& 484   & Image        & Image  & Static & Static Layouts \\

Interaction2Code \cite{xiao2025interaction2code} 
& 127   & Image Pair      & Image  & User Interaction & Interactive Webpages\\

WebVR \cite{dai2026webvr}                        
& 175   & Video        & Video  & User Interaction & Interactive Webpages \\

\midrule
\textbf{Animation2Code} 
& \textbf{1,069} 
& \textbf{Video} 
& \textbf{Video} 
& \textbf{Animation} 
& \textbf{Motion Graphics}\\ 
\bottomrule
\end{tabularx}
\end{table}

\section{The \texttt{Animation2Code} Benchmark}
The \texttt{Animation2Code} benchmark comprises (a) a dataset of 1,069 examples of web animations, each paired with HTML files and inline CSS and/or JavaScript code, (b) a pipeline for deterministically rendering animation code as video, and (c) a set of metrics for measuring the similarity between pairs of animation videos. 
Our benchmark poses the task of \textit{de-rendering}: given a reference video rendered from an example's ground-truth code with our environment, the goal is to map to code which, when rendered with our environment, results in a video as similar to the reference video as possible.
In this section, we describe how we collect the dataset and its resulting statistics, and outline our environment for deterministically rendering video from code.
In Section~\ref{sec:eval}, we describe our metrics and justify their design by measuring agreement with human preference.

\subsection{Task: Animation De-Rendering}
Given a reference video $V_\text{ref}$ that depicts a webpage with dynamic visual behavior, the goal is to generate an executable HTML file $C$ (with inline CSS and JavaScript). Executing $C$ in a browser produces a rendered video $V_\text{gen}$. The objective is to ensure that $V_\text{gen}$ matches $V_\text{ref}$ in both spatial appearance and temporal dynamics.
 
\subsection{Data Curation and Processing} \label{sec:data-curation}
We curate our dataset from 1,069 publicly available CodePen animation examples\footnote{All public CodePen content is MIT-licensed: \url{https://blog.codepen.io/documentation/licensing/}.} spanning diverse CSS and JavaScript animation styles. Raw source files are normalized to plain HTML files with inline CSS and JavaScript using dedicated compilers or GPT-4.1 for less common formats. We remove interactive triggers to obtain clean animations that autoplay on render. Compound examples containing multiple independent animations are split into individual examples. Each example is compiled into a single HTML file and deterministically rendered to MP4 (1024$\times$768, 30 FPS) using a headless Chromium pipeline that steps CSS animations frame-by-frame via the Web Animations API and synchronizes JavaScript-driven animations with deterministic timing control. The final dataset contains 1,069 pairs of HTML code and rendered video. We randomly split the data into 72\% train, 8\% validation, and 20\% test sets, with sub-examples from the same source grouped together to prevent leakage. Details of each processing step are provided in \autoref{app:data}.

\subsection{Dataset Statistics and Diversity}

\paragraph{Code Complexity} Our dataset comprises over 355k total lines of code (min=7, max=10,657, mean=$333\pm828$). Out of the 1,069 examples, 890 (83.3\%) are pure CSS animations, 169 (15.8\%) incorporate JavaScript (150 with inline scripts, 19 relying on an external script), and 10 (0.9\%) animate with SVG alone. The dataset contains 515 unique CSS properties, with a long-tail distribution where 33.4\% appear only once. The most common animation-related properties are \texttt{animation} (81.6\% of total examples) and \texttt{transform} (72.7\%), which specify animation timing and behavior, and geometric transformations, respectively. See Appendix~\ref{app:css_properties} for the full distribution of animation-related properties.

\paragraph{Animation Diversity} The animation durations vary from 2s to 8s (mean=$5.1\pm 2.4$). Looping animations are captured for at least one full cycle. To characterize motion patterns across the dataset, we use GPT-5-mini to automatically annotate the types of motion present in the HTML files. The animations exhibit diverse motion patterns, including rotation (58.9\%), translation (52.3\%), scaling (35.0\%), and appearance changes such as opacity (25.9\%). These motion primitives often co-occur within a single example, with 65.9\% of the examples combining three or more motion patterns, requiring models to capture compositional spatial and temporal dynamics.

\section{Evaluating De-Rendering}
\label{sec:eval}

\begin{figure}[t]
\centering
\includegraphics[width=\linewidth]{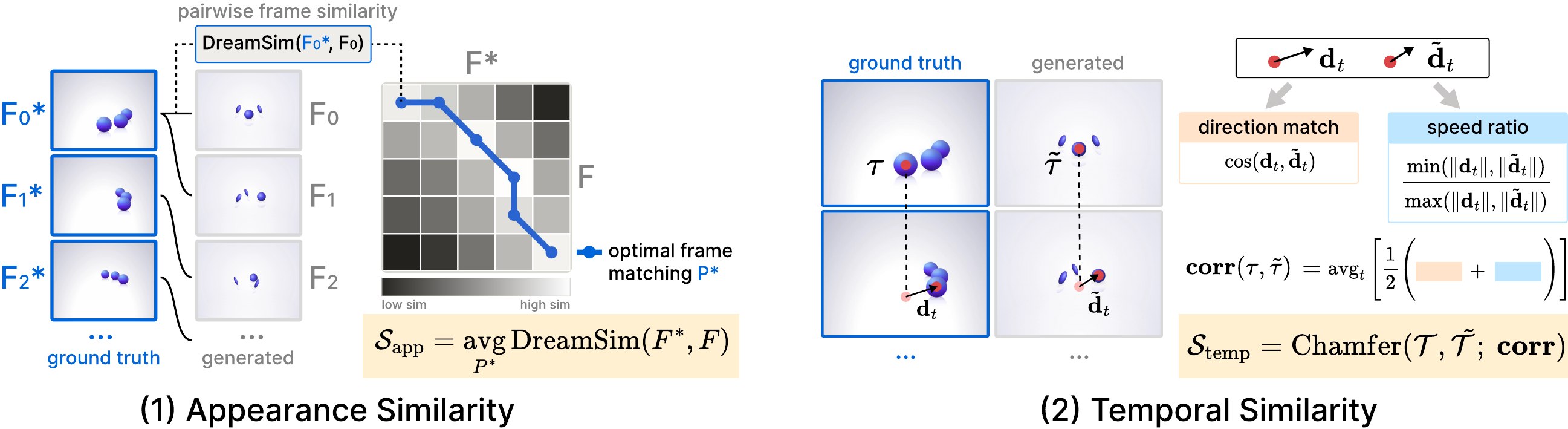}
\caption{Evaluation metrics for \texttt{Animation2Code}. We isolate animated regions via cropping, compute appearance similarity using DreamSim with DTW alignment, and measure temporal similarity via tracklet displacement correlation aggregated with a Chamfer-style matching.}
\label{fig:metrics}
\end{figure}

We evaluate generated animations in two main aspects: \textit{appearance} and \textit{temporal} similarity. These aspects capture complementary qualities: an animation may reproduce the correct visual appearance of elements while failing to match their underlying motion dynamics, or vice versa.

\paragraph{Extracting Frames from Animation Region}
To isolate the measurement of reconstruction of visual appearance and temporal dynamics, we design both metrics to be agnostic to the absolute positioning of an animation within the rendered webpage. 
Absolute position can introduce significant confound in the metrics even when appearance and motion are well matched to the target (\autoref{app:full-frame}). In practice, animated components are usually embedded in a webpage where their absolute position is determined by the surrounding context. As a result, our benchmark evaluates animations in isolation by restricting to the animated area. 
We (1) crop both reference and generated video to its own animated bounding box (the smallest square enclosing all pixels that have an absolute change above a threshold relative to the first frame), (2) downscale the larger crop to match the smaller, and (3) measure similarity based on the animated crops.

\paragraph{Appearance Similarity}
Given the reference animation video frames $V_\text{ref} = \{F^*_i\}^n_{i=1}$ and video frames $V_\text{gen} = \{F_j\}^m_{j=1}$ rendered from the generated code, we only focus on the best match of color, shape, and style of the animated element across video frames.  We measure appearance similarity using DreamSim \cite{fu2023dreamsim} embeddings between frames aligned via Dynamic Time Warping (DTW) \cite{bringmann2023dynamic}. DreamSim is trained using an ensemble of concatenated features from backbone models and specifically tuned to better align with human perception. 
We use the DreamSim distance to calculate the DTW cost for each frame pair $(F^*_i, F_j)$, 
\begin{equation}
    D_{i,j} = 1 - \cos(\phi(F^*_i), \phi(F_j)),
    \label{eq:dtw-cost}
\end{equation}
where $\phi$ is the DreamSim encoder.
DTW finds the optimal alignment path between two videos $P^*=\{(i_l,j_l)\}_{l=1}^{L}$, and the appearance similarity score is the mean similarity along that path:
\begin{equation}
  \mathcal{S}_{\mathrm{Appearance}}
    = \frac{1}{L}\sum_{l=1}^{L}\bigl(1 - D_{i_l,j_l}\bigr).
  \label{eq:as-score}
\end{equation}

\paragraph{Temporal Similarity}
We measure the similarity between the motion trajectories of the ground truth and generated animations, independent of the visual appearance of the animated elements.
Traditional motion metrics, such as comparing optical-flow fields, assume pixel-level correspondence between videos. However, generated animations often differ structurally from the ground truth, making such alignment assumptions invalid. To address this, we build on the Motion Fidelity metric proposed by Yatim et al.~\cite{yatim2024space}, which compares motion trajectories without requiring spatial alignment.

Given a pair of videos, we extract tracklets using CoTracker3~\cite{karaev2025cotracker3}. Since CoTracker3 is trained for tracking in natural videos and does not directly generalize to web animations, we initialize tracking points programmatically. Specifically, we partition each frame into a $30{\times}30$ grid and, for each cell, select the pixel whose value first changes from the initial frame beyond a fixed threshold.
We then query CoTracker3 at the first appearance of each selected point, enabling robust tracking of progressively introduced animations.

This yields tracklet sets $\mathcal{T} = \{\tau_i\}_{i=1}^m$ and $\tilde{\mathcal{T}} = \{\tilde{\tau}_j\}_{j=1}^n$, where $q_i$ and $\tilde{q}_j$ denote the first-appearance frame of each tracklet. Let $\mathbf{d}_k^i = \tau_i(k{+}1) - \tau_i(k)$ be the displacement at step $k$ (zero for $k < q_i$). The correlation between two tracklets is:
\begin{equation}
  \mathrm{\textbf{corr}}(\tau_i,\tilde{\tau}_j) =
    \frac{1}{T - q_{ij}}\sum_{k=q_{ij}}^{T-1}
    \frac{1}{2}\!\left[
      \frac{\mathbf{d}_k^i \cdot \tilde{\mathbf{d}}_k^j}{\|\mathbf{d}_k^i\|\,\|\tilde{\mathbf{d}}_k^j\|}
      \;+\;
      \frac{\min(\|\mathbf{d}_k^i\|,\|\tilde{\mathbf{d}}_k^j\|)}{\max(\|\mathbf{d}_k^i\|,\|\tilde{\mathbf{d}}_k^j\|)}
    \right],
\end{equation}
where $q_{ij} = \min(q_i, \tilde{q}_j)$ is the earlier start frame between two tracklets and $T$ is the total number of steps. The first term measures direction agreement (cosine similarity); the second measures speed agreement (ratio of magnitudes $\in [0,1]$).

The temporal similarity score aggregates pairwise correlations via Chamfer distance:
\begin{equation}
  \mathcal{S}_{\mathrm{Temporal}}
    = \frac{1}{2}\!\left(
        \frac{1}{m}\sum_{\tilde{\tau}\in\tilde{\mathcal{T}}}
          \max_{\tau\in\mathcal{T}}\mathrm{\textbf{corr}}(\tau,\tilde{\tau})
        +
        \frac{1}{n}\sum_{\tau\in\mathcal{T}}
          \max_{\tilde{\tau}\in\tilde{\mathcal{T}}}\mathrm{\textbf{corr}}(\tau,\tilde{\tau})
      \right).
\end{equation}

\subsection{Validating Metrics}
To assess whether our automatic metrics reflect human perception of animation quality, we compare the performance of metrics with human annotations of the zero-shot baseline model outputs. We evaluate both pairwise model preferences and the agreement between human annotations and automatic metrics.

\paragraph{Setup}
We conduct pairwise preference annotation on samples from 10 representative model pairs (Gemini 3 Flash Preview, Qwen3-VL-8B-Instruct, GPT-5.4, Claude Sonnet 4.6, and LLaMA 4 Scout) (exact prompts in Appendix~\ref{app:prompts}), with 60 examples per pair drawn from 214 test set animations under zero-shot generation, yielding 600 comparisons in total. For Gemini and Qwen3-VL, inputs are provided as 2 FPS videos; for the remaining models, we use 2 FPS sampled frames.  
We recruited 65 annotators via Prolific \cite{prolific2026}, paid \$18.00 per hour. Each comparison is independently rated by 3--4 annotators on three dimensions: \emph{overall quality}, \emph{appearance similarity} to the reference, and \emph{temporal similarity} to the reference. Final labels are determined by majority vote.

\begin{figure}[t]
\centering
\includegraphics[width=\linewidth]{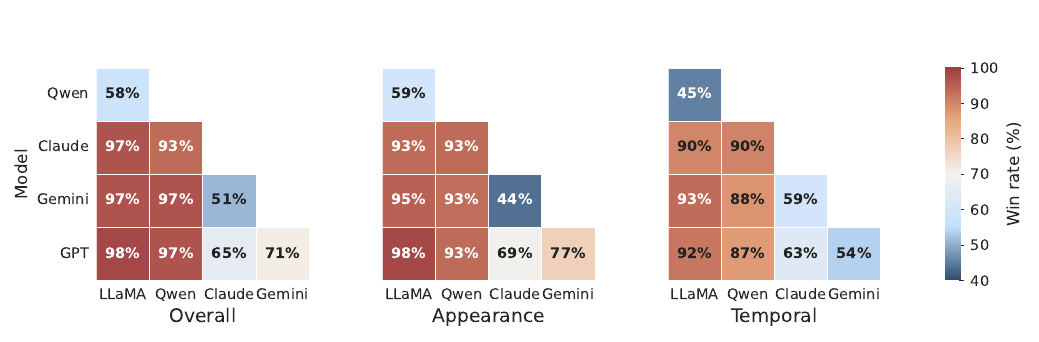}
\caption{Pairwise win rates from human preference annotations (ties excluded). Rows indicate the preferred model over columns. \textbf{GPT-5.4 is the most preferred over all tested models. Models that excel in appearance do not consistently achieve better temporal alignment.}}
\label{fig:human-results}
\end{figure}

\paragraph{Human Preference Results}
\autoref{fig:human-results} shows pairwise win rates based on human annotations. The tie rate is approximately 5.6\% overall (6.0\% for appearance and 9.3\% for temporal similarity). 
GPT-5.4 is consistently preferred across most comparisons, while LLaMA 4 Scout performs worst overall, which aligns with the results from automatic metrics. Temporal similarity shows relatively flatter win rates and higher tie rates, suggesting that current models are less differentiated in their ability to reproduce motion dynamics. 
Notably, better quality in appearance does not consistently translate to better temporal alignment, suggesting recovering temporal fidelity is harder.

\begin{figure}[t]
\centering
\includegraphics[width=\linewidth]{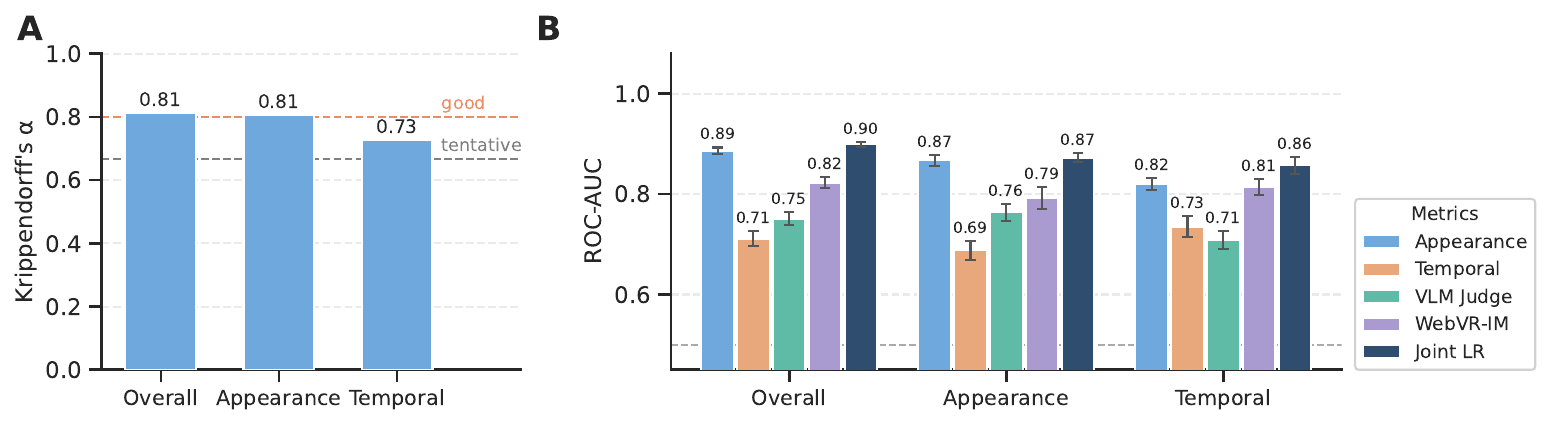}
\caption{(A) Inter-annotator agreement (Krippendorff's $\alpha$, ties excluded).  Dashed lines mark ``tentative'' ($\alpha=0.67$) and ``good'' ($\alpha=0.80$) agreement thresholds \cite{krippendorff2018content}. (B) ROC-AUC \cite{bradley1997use} for predicting pairwise human preference. Joint LR combines both metric deltas using logistic regression trained on a 50\% subset of human annotations. Error bars indicate 95\% CI over 10 random splits. \textbf{Human annotations are reliable but noisier for temporal similarity. Joint LR aligns best with human preference in every dimension.}}
\label{fig:agreement}
\end{figure}

We measure the reliability of human annotations with Krippendorff's $\alpha$~\cite{krippendorff2011computing,krippendorff2018content}. The result in \autoref{fig:agreement}A indicates strong inter-annotator agreement for overall (0.81) and appearance (0.81) annotations. Temporal annotations achieve lower agreement (0.73) but remain above the threshold for tentative conclusions (0.67), indicating that motion quality is inherently harder to judge than appearance~\cite{tversky2002animation,kaiser1992influence,betrancourt2005animation}, while still providing a consistent signal for evaluation.

Overall annotations correlate more strongly with appearance than temporal (Spearman’s $\rho=0.82$ vs. $\rho=0.73$, $p\ll0.001$). Appearance and temporal annotations show only moderate correlation ($\rho=0.60$, $p\ll0.001$), indicating that both capture complementary aspects of animation quality.

\paragraph{Alignment with Human Judgments}
We validate our appearance and temporal similarity metrics against the human annotations using the area under the ROC curve (ROC-AUC) \cite{bradley1997use}. For each pairwise comparison, we compute the difference in the metric scores to predict the binary human judgment. We compare against two VLM-based baselines using Gemini 3.1 Pro: a VLM judge prompted with the same task as human annotators (Appendix~\ref{app:vlm-judge-prompt}) and the Interaction-and-Motion (IM) metric adapted from the WebVR benchmark \cite{dai2026webvr} (Appendix~\ref{app:webvr-im}). 
As shown in \autoref{fig:agreement}B, the appearance metric achieves strong agreement with human preferences, matching or outperforming both VLM-based baselines across all three dimensions. 
While the temporal metric achieves lower standalone agreement, it aligns most strongly with human judgments of temporal similarity than with judgments of the other two dimensions. 

To examine whether the metrics capture complementary information, we fit a logistic regression model to predict human preference using both appearance and temporal score differences as features, with a 50/50 train/test split of the human pairwise annotation data by animation. Both features are significant predictors ($p<0.001$; except for temporal score for appearance with $p<0.05$). Combining both metrics with logistic regression (Joint LR) achieves best agreement with human annotations across all dimensions in terms of ROC-AUC, with corresponding test accuracies of 83.9\%, 82.2\%, and 81.2\% for overall, appearance, and temporal judgments, respectively, averaged over 10 random splits.
These results demonstrate that appearance and temporal similarity capture complementary aspects of animation quality. Together, they provide a human-aligned evaluation that separately measures visual appearance and temporal fidelity in animation reconstruction, without relying on VLM-based judging.

\section{Benchmarking: Prompting, Fine-tuning, and Refinement} \label{sec:benchmarking}

\subsection{Experimental Setup}

\paragraph{Zero-shot Baselines}
We evaluate a mix of commercial and open-source multimodal models under a zero-shot setting with direct prompting (exact prompts in Appendix~\ref{app:prompts}): Gemini 3 Flash Preview, Qwen3-VL-8B-Instruct, GPT-5.4, Claude Sonnet 4.6, and LLaMA 4 Scout. Each model receives a short instruction along with the target animation and is asked to return a single HTML document.

We consider two input modalities: (1) native video input, when supported by the model, and (2) image-frame input, where the video is uniformly subsampled into frames before being passed to the model. For Gemini and Qwen3-VL, which both accept native video, we report three input FPS settings: 2 FPS and the best input FPS for each metric based on a full FPS sweep from 0.5 to 24 FPS (Appendix~\ref{app:fps}). For the remaining models, we use frames sampled at 2 FPS.

\paragraph{Supervised Fine-tuning}
We fine-tune Qwen3-VL-8B-Instruct on the animation-to-code task using LoRA \citep{hu2022lora} and full-parameter fine-tuning. Both use the same training data, preprocessing, and inference setup. Full hyperparameters and infrastructure details are provided in Appendix~\ref{app:sft-details}.

\paragraph{Iterative Refinement}
To improve generation quality beyond a single pass, we adopt an iterative refinement procedure in which the model repeatedly compares its rendered output against the target animation and revises the code. 
We adapt METAL \citep{li2025metal}, a multi-agent framework for iterative chart refinement, to our task. The framework uses two critics: a visual critic operating on rendered outputs and a code critic operating on program text. The editor agent updates the program based on both signals, enabling correction of both visual and code-level errors. We use Qwen3-VL-8B-Instruct as the backbone for all agents and run 3 iterations. We additionally compare against three alternative refinement variants in Appendix~\ref{app:ir-variants}.
\subsection{Results and Analysis}

\begin{table*}[t]
\centering
\small
\caption{Test set zero-shot, supervised fine-tuning, and iterative refinement performance across models and input settings. Best score under each metric is highlighted in \textbf{bold}. Exec\% denotes the percentage of examples where the generated code executes successfully. \textbf{While some models achieve near-perfect execution and strong appearance similarity, temporal similarity remains uniformly low across settings, highlighting a consistent gap in replicating animation dynamics.} }
\begin{tabular}{l l c c c c}
\toprule
\textbf{Setting} & \textbf{Model} & \textbf{FPS} & \textbf{Exec (\%)} & \textbf{Appearance} & \textbf{Temporal} \\
\midrule

\multirow{6}{*}{\centering Native video input}
& Gemini 3 Flash Preview       & 2  & 98.1 & 0.80 & 0.30 \\
& Gemini 3 Flash Preview       & 4  & 98.6 & 0.81 & 0.31 \\
& Gemini 3 Flash Preview       & 16 & 99.5 & 0.80 & \textbf{0.32} \\
& Qwen3-VL-8B-Instruct         & 2  & 85.5 & 0.67 & 0.23 \\
& Qwen3-VL-8B-Instruct         & 4  & 82.2 & 0.68 & 0.25 \\
& Qwen3-VL-8B-Instruct         & 24 & 84.6 & 0.69 & 0.24 \\

\midrule

\multirow{5}{*}{\centering Image frames input}
& GPT-5.4                      & 2  & \textbf{100.0} & \textbf{0.84} & 0.29 \\
& Gemini 3 Flash Preview       & 2  & \textbf{100.0} & 0.80 & 0.30 \\
& Claude Sonnet 4.6            & 2  & 99.5 & 0.82  & 0.29 \\
& LLaMA 4 Scout                & 2  & 97.7 & 0.62 & 0.21 \\
& Qwen3-VL-8B-Instruct         & 2  & 80.4 & 0.70 & 0.24 \\

\midrule

\multirow{4}{*}{\parbox[c]{3cm}{SFT / refinement\\(Video input)}}
& \textit{Qwen3-VL-8B-Instruct}        &   &  &  & \\
& LoRA                       & 2  & 88.3 & 0.56 & 0.22 \\
& Full SFT                   & 2  & 87.4 & 0.56 & 0.21 \\
& Iterative Refinement       & 2  & 85.5 & 0.73 & 0.28 \\

\bottomrule
\end{tabular}
\label{tab:results}
\end{table*}

\paragraph{Appearance and Temporal Similarity}
Execution success is nearly saturated for most models ($\geq$97\%), and appearance similarity is also relatively high, with GPT-5.4 achieving the best score (0.84), suggesting strong capability in reproducing static visual layout.
In contrast, temporal similarity remains uniformly low across models. Gemini achieves the highest temporal score (0.32) at its best input frame rate of 16 FPS. Notably, high appearance scores do not translate to strong temporal performance.

Both fine-tuned Qwen3-VL-8B-Instruct models improve execution success by learning the correct HTML structure, but degrade appearance and temporal similarity. 
This is because supervised fine-tuning optimizes a code-level objective, but that does not necessarily carry over to the visual reconstruction quality. The fine-tuned models more often declare animation in code that did not end up being visible in the render, with static renders rising from 4.4\% (zero-shot) to 9.0\% (LoRA) and 11.8\% (full SFT). This reveals a gap between executable code generation and modeling continuous temporal dynamics.

Iterative self-refinement based on Qwen3-VL-8B-Instruct's zero-shot baseline output effectively improves both appearance and temporal similarity over three iterations. The largest gains occur in the first iteration (+4.4\% appearance and +9.0\% temporal similarity), followed by diminishing improvements in later iterations (+0.9\% appearance and +3.0\% temporal similarity in the final iteration).

\paragraph{Input FPS} Across input settings, native video input provides only modest gains over image-frame input, and increasing frame rate yields only marginal improvement in both appearance and temporal scores. 
Over a full 0.5 to 24 FPS sweep (Appendix~\ref{app:fps}), appearance similarity plateaus early with 0.80 at 4 FPS for Gemini but improves consistently for Qwen from 0.66 to 0.69. For temporal similarity, Gemini improves by at most 0.04 (0.5 to 16 FPS), while Qwen does not show a clear trend. This suggests that simply providing richer temporal signals is insufficient for better modeling continuous animation.

\subsection{Failure Analysis}
\begin{figure}[t]
\centering
\includegraphics[width=\linewidth]{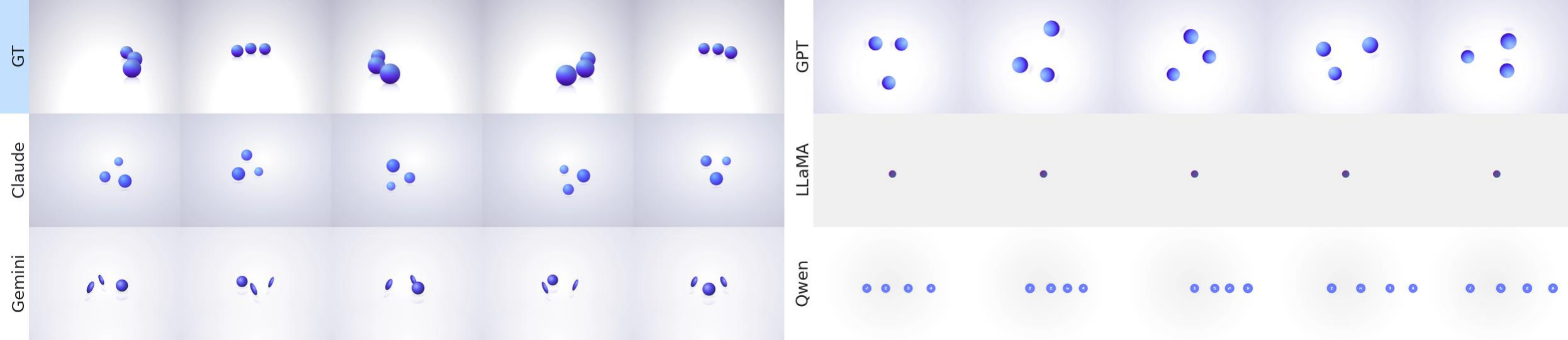}
\includegraphics[width=\linewidth]{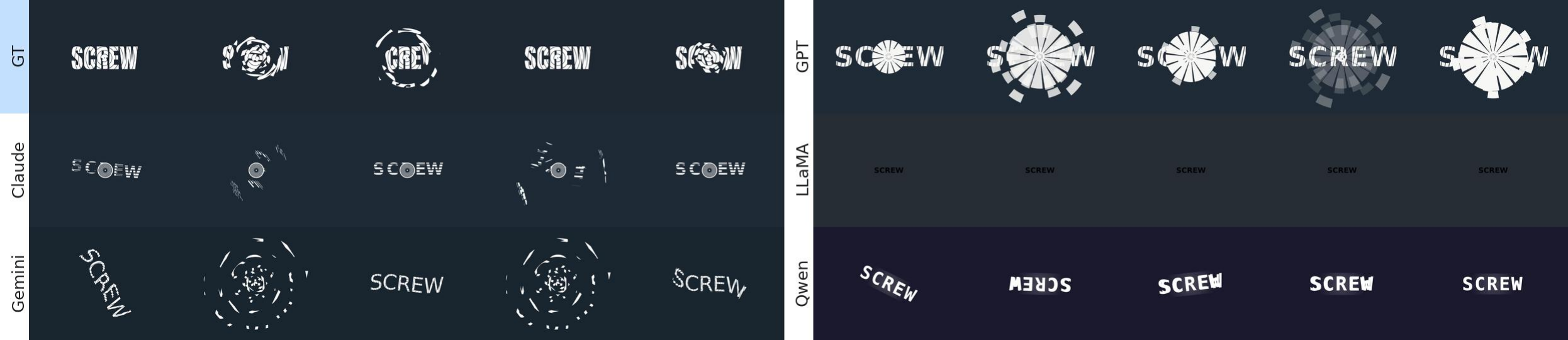}
\includegraphics[width=\linewidth]{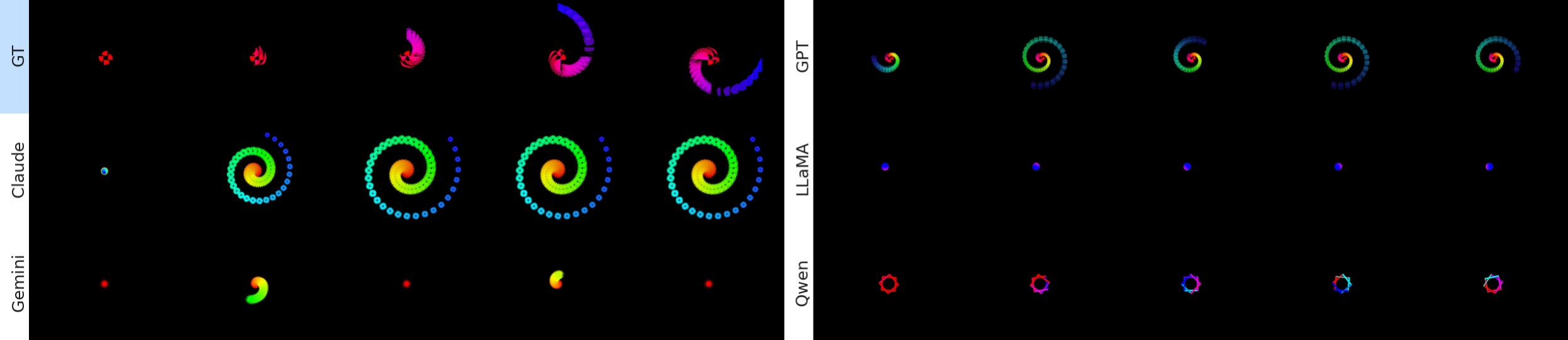}
\caption{Qualitative examples from zero-shot baseline models. \textbf{All models struggle with precise motion and spatial relationships, even when overall appearance is partially correct.}}
\label{fig:example}
\end{figure}

\begin{figure}[t]
\centering
\includegraphics[width=\linewidth]{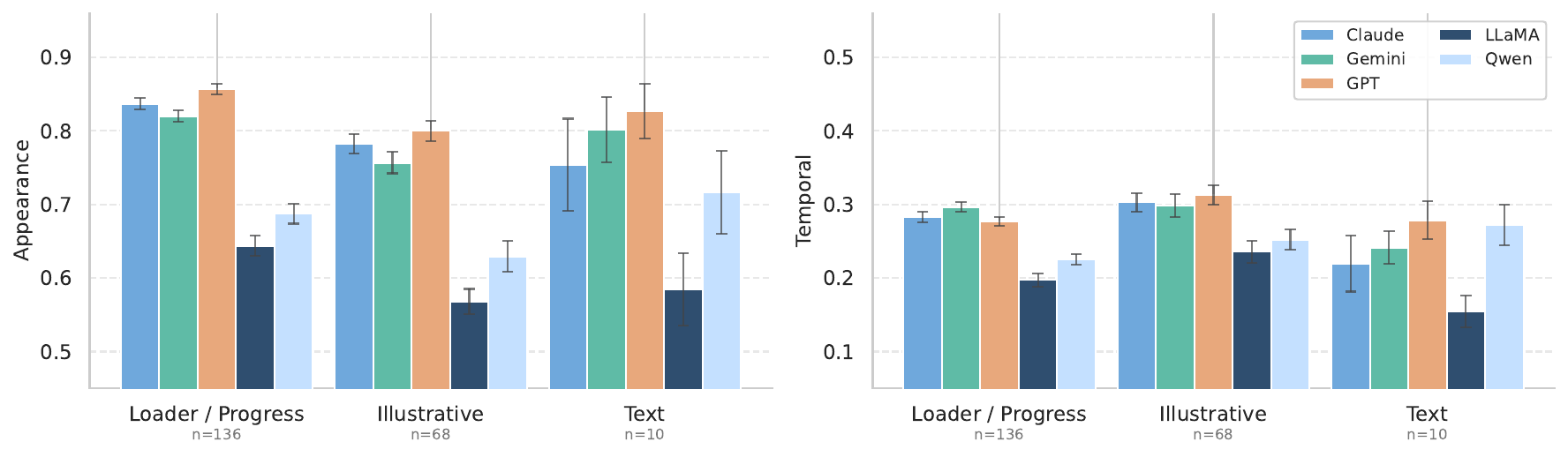}
\caption{Baseline performance by animation category. \textbf{Appearance varies with visual complexity, but temporal performance remains consistently low across categories.}}
\label{fig:category}
\end{figure}

We provide further qualitative analysis of rendered results on the test set using the main zero-shot baselines: Gemini and Qwen with 2 FPS video input, and GPT, Claude, and LLaMA with 2 FPS sampled image frames.
\textbf{(1) Qualitative comparison: }
In \autoref{fig:example}, GPT and Claude best capture the appearance of 3D balls, while LLaMA and Qwen render them as 2D circles. GPT, Claude, and Gemini replicate the circling motion but fail to reproduce the correct movement trajectory and spatial arrangement, whereas LLaMA produces static visuals and Qwen produces linear motion.
\textbf{(2) Performance by category:} 
\autoref{fig:category} shows the performance of each model by animation types categorized using GPT-5-mini based on the ground-truth code. Models achieve lower appearance fidelity on illustrative animations (e.g., objects and abstract geometric visuals), likely due to their higher complexity compared to loader/progress or text animations, while temporal similarity remains relatively uniform across categories.
\textbf{(3) Static rate:}
While models generate executable code, not all render into animation. A static render is scored 0 on temporal similarity to the ground truth. GPT and Gemini achieve the 0\% static rate, followed by Claude (0.5\%), Qwen (4.4\%), and LLaMA (12.0\%).

\section{Related Work}
\paragraph{Temporal Reasoning with Videos}
General-purpose video benchmarks \cite{fu2025video,li2024mvbench,li2024videovista, ning2025video} evaluate VLMs on real-world videos that require temporal reasoning over action sequences, object interactions, and event dynamics, revealing consistent weaknesses in temporal understanding. 
More specialized benchmarks probe finer-grained temporal understanding. TempCompass \cite{liu2024tempcompass} isolates temporal perception across action, speed, direction, and attribute changes. TemporalBench \cite{cai2024temporalbench} tests fine-grained temporal dynamics such as action frequency, motion magnitude, and event order. TimeBlind \cite{li2026timeblind} demonstrates the brittleness of frontier models in distinguishing temporal dynamics using video pairs controlled for static visual content, where motion is the only difference. 
A parallel line of work uses synthetic videos to diagnose temporal reasoning in controlled settings. CLEVRER \cite{yi2019clevrer} introduced causal and temporal reasoning over simple 3D objects rendered in Blender, and SynRL \cite{jiang2026learning} recently showed that temporal primitives learned from programmatically generated videos of geometric shapes transfer to real-world understanding. 
Across both real-world and synthetic settings, existing benchmarks evaluate temporal reasoning in videos as a pure comprehension task. Our benchmark requires models to not only reason about temporal dynamics but also produce executable code that reproduces the observed behavior.

\paragraph{Visually Grounded Code Synthesis}
The task of visually grounded code synthesis (also referred to as de-rendering) has been studied primarily in static settings, where models generate executable code from image input.
Prior work spans domains such as charts and plots \cite{wu2025plot2code, niu2025chart2code53, zhao2025chartcoder}, webpage screenshots \cite{beltramelli2018pix2code, laurenccon2024unlocking, si2025design2code, xiao2025interaction2code, roberts2024image2struct,yun2024web2code}, presentation slides \cite{ge2025autopresent, yang2026slidesgen}, SVG graphics \cite{zou2024vgbench,li2025unisvg}, LaTeX expressions \cite{deng2017image, roberts2024image2struct}, and music scores \cite{roberts2024image2struct}. 
These works focus on recovering spatial structure and layout from static images. De-rendering dynamic visuals introduces additional challenges: models must infer how visual elements evolve over time and produce code that reproduces the observed temporal behavior. A recent benchmark \cite{dai2026webvr} evaluates VLMs on recreating webpages from demonstration videos, but it mainly targets interactive behaviors and does not systematically evaluate complex animated visual elements. We distinguish animation, i.e., continuous, time-driven visual change independent of user input, from interaction, which is discrete state changes triggered by user events (clicks, hovers, scrolls). Existing efforts in code synthesis for animation primarily focus on interactive systems that assist users in creating animations from static images and text instructions \cite{qiu2025anyani, liu2025logomotion, park2025decomate}, rather than evaluating models on inferring animation logic from visual observations.
To the best of our knowledge, no benchmark exists for visually grounded code synthesis of dynamic web animations.

\section{Conclusion} \label{sec:conclusion}
We introduced \texttt{Animation2Code}, the first benchmark for evaluating temporal visual reasoning in video-to-code generation through the task of animation de-rendering. We also proposed a human-aligned evaluation suite that disentangles appearance and temporal fidelity of generated animations. 
Across zero-shot prompting, supervised fine-tuning, and iterative refinement, we find that while current models can reliably produce executable code and achieve strong appearance similarity, they consistently fail to capture correct temporal dynamics. 

These results highlight a fundamental gap in temporal visual reasoning in state-of-the-art VLMs through video-to-code de-rendering. \texttt{Animation2Code} provides a controlled testbed and human-aligned metrics for diagnosing and improving temporal visual reasoning in VLMs, isolating challenges in recovering temporal dynamics that are otherwise ignored or confounded with appearance mismatches in existing benchmarks.
We hope the benchmark will support future research in multimodal program synthesis and temporal visual reasoning, while encouraging responsible use by mitigating risks such as unlicensed cloning of web visuals through MIT-licensed sourcing and preserved attribution.
\section{Limitations} \label{sec:limitations}

Our temporal similarity metric relies on tracklets extracted with CoTracker3, which is not optimized for synthetic web animations. As a result, tracking can be noisy or unstable for small, fast, or low-texture elements, introducing noise into the motion similarity estimate. While the metric shows consistent alignment with human judgments, these imperfections may limit its ability to capture fine-grained temporal dynamics.

Our benchmark focuses on web animations derived from CodePen, which, while diverse, may not cover all forms of dynamic visual content (e.g., animated mathematical visualizations like Manim, physics-based 3D simulations \cite{liang2026visphyworld,xie2026physcodebench,lu2026worldcoder}, or animations that use specialized libraries). Extending the benchmark to broader domains and exploring more advanced optimization methods such as RL remain important directions for future work.
\begin{ack}
We thank Haiwen Feng, Téa Wright, Kalvin Chang, and Wenjie Ma for valuable suggestions and feedback, the Prolific workers for participating in the human study, and the CodePen contributors whose publicly available projects were used in constructing our benchmark.
This research was supported by an Amazon Research Award, a gift from Google, a Technical AI Safety Research award from Coefficient Giving, and an NVIDIA Academic Grant Program award.
\end{ack}



\bibliography{reference}

@inproceedings{wu2025plot2code,
  title={Plot2code: A comprehensive benchmark for evaluating multi-modal large language models in code generation from scientific plots},
  author={Wu, Chengyue and Liang, Zhixuan and Ge, Yixiao and Guo, Qiushan and Lu, Zeyu and Wang, Jiahao and Shan, Ying and Luo, Ping},
  booktitle={Findings of the Association for Computational Linguistics: NAACL 2025},
  pages={3006--3028},
  year={2025}
}

@inproceedings{niu2025chart2code53,
  title={Chart2Code53: A Large-Scale Diverse and Complex Dataset for Enhancing Chart-to-Code Generation},
  author={Niu, Tianhao and Cui, Yiming and Wang, Baoxin and Xu, Xiao and Yao, Xin and Zhu, Qingfu and Wu, Dayong and Wang, Shijin and Che, Wanxiang},
  booktitle={Proceedings of the 2025 Conference on Empirical Methods in Natural Language Processing},
  pages={15839--15855},
  year={2025}
}

@inproceedings{zhao2025chartcoder,
  title={Chartcoder: Advancing multimodal large language model for chart-to-code generation},
  author={Zhao, Xuanle and Luo, Xianzhen and Shi, Qi and Chen, Chi and Wang, Shuo and Liu, Zhiyuan and Sun, Maosong},
  booktitle={Proceedings of the 63rd Annual Meeting of the Association for Computational Linguistics (Volume 1: Long Papers)},
  pages={7333--7348},
  year={2025}
}

@inproceedings{beltramelli2018pix2code,
  title={pix2code: Generating code from a graphical user interface screenshot},
  author={Beltramelli, Tony},
  booktitle={Proceedings of the ACM SIGCHI symposium on engineering interactive computing systems},
  pages={1--6},
  year={2018}
}

@inproceedings{si2025design2code,
  title={Design2code: Benchmarking multimodal code generation for automated front-end engineering},
  author={Si, Chenglei and Zhang, Yanzhe and Li, Ryan and Yang, Zhengyuan and Liu, Ruibo and Yang, Diyi},
  booktitle={Proceedings of the 2025 Conference of the Nations of the Americas Chapter of the Association for Computational Linguistics: Human Language Technologies (Volume 1: Long Papers)},
  pages={3956--3974},
  year={2025}
}

@article{laurenccon2024unlocking,
  title={Unlocking the conversion of web screenshots into html code with the websight dataset},
  author={Lauren{\c{c}}on, Hugo and Tronchon, L{\'e}o and Sanh, Victor},
  journal={arXiv preprint arXiv:2403.09029},
  year={2024}
}

@inproceedings{xiao2025interaction2code,
  title={Interaction2code: Benchmarking mllm-based interactive webpage code generation from interactive prototyping},
  author={Xiao, Jingyu and Wan, Yuxuan and Huo, Yintong and Wang, Zixin and Xu, Xinyi and Wang, Wenxuan and Xu, Zhiyao and Wang, Yuhang and Lyu, Michael R},
  booktitle={2025 40th IEEE/ACM International Conference on Automated Software Engineering (ASE)},
  pages={241--253},
  year={2025},
  organization={IEEE}
}

@inproceedings{ge2025autopresent,
  title={Autopresent: Designing structured visuals from scratch},
  author={Ge, Jiaxin and Wang, Zora Zhiruo and Zhou, Xuhui and Peng, Yi-Hao and Subramanian, Sanjay and Tan, Qinyue and Sap, Maarten and Suhr, Alane and Fried, Daniel and Neubig, Graham and others},
  booktitle={Proceedings of the Computer Vision and Pattern Recognition Conference},
  pages={2902--2911},
  year={2025}
}

@article{roberts2024image2struct,
  title={Image2struct: Benchmarking structure extraction for vision-language models},
  author={Roberts, Josselin S and Lee, Tony and Wong, Chi H and Yasunaga, Michihiro and Mai, Yifan and Liang, Percy},
  journal={Advances in Neural Information Processing Systems},
  volume={37},
  pages={115058--115097},
  year={2024}
}

@inproceedings{deng2017image,
  title={Image-to-markup generation with coarse-to-fine attention},
  author={Deng, Yuntian and Kanervisto, Anssi and Ling, Jeffrey and Rush, Alexander M},
  booktitle={International Conference on Machine Learning},
  pages={980--989},
  year={2017},
  organization={PMLR}
}

@article{qiu2025anyani,
  title={AnyAni: An Interactive System with Generative AI for Animation Effect Creation and Code Understanding in Web Development},
  author={Qiu, Tianrun and Ma, Yuxin},
  journal={arXiv preprint arXiv:2506.21962},
  year={2025}
}

@inproceedings{liu2025logomotion,
  title={Logomotion: Visually-grounded code synthesis for creating and editing animation},
  author={Liu, Vivian and Kazi, Rubaiat Habib and Wei, Li-Yi and Fisher, Matthew and Langlois, Timothy and Walker, Seth and Chilton, Lydia},
  booktitle={Proceedings of the 2025 CHI Conference on Human Factors in Computing Systems},
  pages={1--16},
  year={2025}
}

@article{park2025decomate,
  title={Decomate: Leveraging Generative Models for Co-Creative SVG Animation},
  author={Park, Jihyeon and Myung, Jiyoon and Shin, Seone and Son, Jungki and Han, Joohyung},
  journal={arXiv preprint arXiv:2511.06297},
  year={2025}
}

@article{yang2026slidesgen,
  title={SlidesGen-Bench: Evaluating Slides Generation via Computational and Quantitative Metrics},
  author={Yang, Yunqiao and Li, Wenbo and Ren, Houxing and Lu, Zimu and Wang, Ke and Huang, Zhiyuan and Zong, Zhuofan and Zhan, Mingjie and Li, Hongsheng},
  journal={arXiv preprint arXiv:2601.09487},
  year={2026}
}

@inproceedings{zou2024vgbench,
  title={Vgbench: Evaluating large language models on vector graphics understanding and generation},
  author={Zou, Bocheng and Cai, Mu and Zhang, Jianrui and Lee, Yong Jae},
  booktitle={Proceedings of the 2024 Conference on Empirical Methods in Natural Language Processing},
  pages={3647--3659},
  year={2024}
}

@inproceedings{li2025unisvg,
  title={Unisvg: A unified dataset for vector graphic understanding and generation with multimodal large language models},
  author={Li, Jinke and Yu, Jiarui and Wei, Chenxing and Dong, Hande and Lin, Qiang and Yang, Liangjing and Wang, Zhicai and Hao, Yanbin},
  booktitle={Proceedings of the 33rd ACM International Conference on Multimedia},
  pages={13156--13163},
  year={2025}
}

@article{dai2026webvr,
  title={WebVR: Benchmarking Multimodal LLMs for WebPage Recreation from Videos via Human-Aligned Visual Rubrics},
  author={Dai, Yuhong and Lai, Yanlin and Huang, Mitt and Guo, Hangyu and Li, Dingming and Peng, Hongbo and Li, Haodong and Zhao, Yingxiu and Lyu, Haoran and Ge, Zheng and others},
  journal={arXiv preprint arXiv:2603.13391},
  year={2026}
}

@article{jiang2026learning,
  title={Learning Transferable Temporal Primitives for Video Reasoning via Synthetic Videos},
  author={Jiang, Songtao and Song, Sibo and Zhou, Chenyi and Wang, Yuan and Chen, Ruizhe and Guan, Tongkun and Luo, Ruilin and Zhang, Yan and Tang, Zhihang and Sun, Yuchong and others},
  journal={arXiv preprint arXiv:2603.17693},
  year={2026}
}

@inproceedings{fu2025video,
  title={Video-mme: The first-ever comprehensive evaluation benchmark of multi-modal llms in video analysis},
  author={Fu, Chaoyou and Dai, Yuhan and Luo, Yongdong and Li, Lei and Ren, Shuhuai and Zhang, Renrui and Wang, Zihan and Zhou, Chenyu and Shen, Yunhang and Zhang, Mengdan and others},
  booktitle={Proceedings of the IEEE/CVF conference on computer vision and pattern recognition},
  pages={24108--24118},
  year={2025}
}

@inproceedings{li2024mvbench,
  title={Mvbench: A comprehensive multi-modal video understanding benchmark},
  author={Li, Kunchang and Wang, Yali and He, Yinan and Li, Yizhuo and Wang, Yi and Liu, Yi and Wang, Zun and Xu, Jilan and Chen, Guo and Luo, Ping and others},
  booktitle={Proceedings of the IEEE/CVF Conference on Computer Vision and Pattern Recognition},
  pages={22195--22206},
  year={2024}
}

@article{li2024videovista,
  title={Videovista: A versatile benchmark for video understanding and reasoning},
  author={Li, Yunxin and Chen, Xinyu and Hu, Baotian and Wang, Longyue and Shi, Haoyuan and Zhang, Min},
  journal={arXiv preprint arXiv:2406.11303},
  year={2024}
}

@article{ning2025video,
  title={Video-bench: A comprehensive benchmark and toolkit for evaluating video-based large language models},
  author={Ning, Munan and Zhu, Bin and Xie, Yujia and Lin, Bin and Cui, Jiaxi and Yuan, Lu and Chen, Dongdong and Yuan, Li},
  journal={Computational Visual Media},
  year={2025},
  publisher={TUP}
}

@inproceedings{liu2024tempcompass,
  title={Tempcompass: Do video llms really understand videos?},
  author={Liu, Yuanxin and Li, Shicheng and Liu, Yi and Wang, Yuxiang and Ren, Shuhuai and Li, Lei and Chen, Sishuo and Sun, Xu and Hou, Lu},
  booktitle={Findings of the Association for Computational Linguistics: ACL 2024},
  pages={8731--8772},
  year={2024}
}

@article{cai2024temporalbench,
  title={Temporalbench: Benchmarking fine-grained temporal understanding for multimodal video models},
  author={Cai, Mu and Tan, Reuben and Zhang, Jianrui and Zou, Bocheng and Zhang, Kai and Yao, Feng and Zhu, Fangrui and Gu, Jing and Zhong, Yiwu and Shang, Yuzhang and others},
  journal={arXiv preprint arXiv:2410.10818},
  year={2024}
}

@article{li2026timeblind,
  title={TimeBlind: A Spatio-Temporal Compositionality Benchmark for Video LLMs},
  author={Li, Baiqi and Zhao, Kangyi and Zhang, Ce and Mitra, Chancharik and Nyandwi, Jean de Dieu and Bertasius, Gedas},
  journal={arXiv preprint arXiv:2602.00288},
  year={2026}
}

@article{yi2019clevrer,
  title={Clevrer: Collision events for video representation and reasoning},
  author={Yi, Kexin and Gan, Chuang and Li, Yunzhu and Kohli, Pushmeet and Wu, Jiajun and Torralba, Antonio and Tenenbaum, Joshua B},
  journal={arXiv preprint arXiv:1910.01442},
  year={2019}
}

@article{yun2024web2code,
  title={Web2code: A large-scale webpage-to-code dataset and evaluation framework for multimodal llms},
  author={Yun, Sukmin and Lin, Haokun and Thushara, Rusiru and Bhat, Mohammad Q and Wang, Yongxin and Jiang, Zutao and Deng, Mingkai and Wang, Jinhong and Tao, Tianhua and Li, Junbo and others},
  journal={Advances in neural information processing systems},
  volume={37},
  pages={112134--112157},
  year={2024}
}

@article{achiam2023gpt,
  title={Gpt-4 technical report},
  author={Achiam, Josh and Adler, Steven and Agarwal, Sandhini and Ahmad, Lama and Akkaya, Ilge and Aleman, Florencia Leoni and Almeida, Diogo and Altenschmidt, Janko and Altman, Sam and Anadkat, Shyamal and others},
  journal={arXiv preprint arXiv:2303.08774},
  year={2023}
}

@article{fu2023dreamsim,
  title={Dreamsim: Learning new dimensions of human visual similarity using synthetic data},
  author={Fu, Stephanie and Tamir, Netanel and Sundaram, Shobhita and Chai, Lucy and Zhang, Richard and Dekel, Tali and Isola, Phillip},
  journal={arXiv preprint arXiv:2306.09344},
  year={2023}
}

@article{bringmann2023dynamic,
  title={Dynamic dynamic time warping},
  author={Bringmann, Karl and Fischer, Nick and van der Hoog, Ivor and Kipouridis, Evangelos and Kociumaka, Tomasz and Rotenberg, Eva},
  journal={arXiv preprint arXiv:2310.18128},
  year={2023}
}

@inproceedings{yatim2024space,
  title={Space-time diffusion features for zero-shot text-driven motion transfer},
  author={Yatim, Danah and Fridman, Rafail and Bar-Tal, Omer and Kasten, Yoni and Dekel, Tali},
  booktitle={Proceedings of the IEEE/CVF Conference on Computer Vision and Pattern Recognition},
  pages={8466--8476},
  year={2024}
}

@inproceedings{karaev2025cotracker3,
  title={Cotracker3: Simpler and better point tracking by pseudo-labelling real videos},
  author={Karaev, Nikita and Makarov, Yuri and Wang, Jianyuan and Neverova, Natalia and Vedaldi, Andrea and Rupprecht, Christian},
  booktitle={Proceedings of the IEEE/CVF International Conference on Computer Vision},
  pages={6013--6022},
  year={2025}
}

@book{krippendorff2018content,
  title={Content analysis: An introduction to its methodology},
  author={Krippendorff, Klaus},
  year={2018},
  publisher={Sage publications},
  page={241}
}

@article{krippendorff2011computing,
  title={Computing Krippendorff's alpha-reliability},
  author={Krippendorff, Klaus},
  year={2011}
}

@article{bradley1997use,
  title={The use of the area under the ROC curve in the evaluation of machine learning algorithms},
  author={Bradley, Andrew P},
  journal={Pattern recognition},
  volume={30},
  number={7},
  pages={1145--1159},
  year={1997},
  publisher={Elsevier}
}

@misc{prolific2026,
  author       = {{Prolific}},
  title        = {Prolific: Online Participant Recruitment Platform},
  year         = {2026},
  howpublished = {\url{https://www.prolific.com}},
}

@article{hu2022lora,
  title={Lora: Low-rank adaptation of large language models.},
  author={Hu, Edward J and Shen, Yelong and Wallis, Phillip and Allen-Zhu, Zeyuan and Li, Yuanzhi and Wang, Shean and Wang, Liang and Chen, Weizhu and others},
  journal={Iclr},
  volume={1},
  number={2},
  pages={3},
  year={2022}
}

@article{hsu2024liger,
  title={Liger kernel: Efficient triton kernels for llm training},
  author={Hsu, Pin-Lun and Dai, Yun and Kothapalli, Vignesh and Song, Qingquan and Tang, Shao and Zhu, Siyu and Shimizu, Steven and Sahni, Shivam and Ning, Haowen and Chen, Yanning},
  journal={arXiv preprint arXiv:2410.10989},
  year={2024}
}

@article{madaan2023self,
  title={Self-refine: Iterative refinement with self-feedback},
  author={Madaan, Aman and Tandon, Niket and Gupta, Prakhar and Hallinan, Skyler and Gao, Luyu and Wiegreffe, Sarah and Alon, Uri and Dziri, Nouha and Prabhumoye, Shrimai and Yang, Yiming and others},
  journal={Advances in neural information processing systems},
  volume={36},
  pages={46534--46594},
  year={2023}
}

@article{xu2025improved,
  title={Improved Iterative Refinement for Chart-to-Code Generation via Structured Instruction},
  author={Xu, Chengzhi and Wang, Yuyang and Wei, Lai and Sun, Lichao and Huang, Weiran},
  journal={arXiv preprint arXiv:2506.14837},
  year={2025}
}

@inproceedings{li2025metal,
  title={Metal: A multi-agent framework for chart generation with test-time scaling},
  author={Li, Bingxuan and Wang, Yiwei and Gu, Jiuxiang and Chang, Kai-Wei and Peng, Nanyun},
  booktitle={Proceedings of the 63rd Annual Meeting of the Association for Computational Linguistics (Volume 1: Long Papers)},
  pages={30054--30069},
  year={2025}
}

@inproceedings{xue2025phyt2v,
  title={Phyt2v: Llm-guided iterative self-refinement for physics-grounded text-to-video generation},
  author={Xue, Qiyao and Yin, Xiangyu and Yang, Boyuan and Gao, Wei},
  booktitle={Proceedings of the Computer Vision and Pattern Recognition Conference},
  pages={18826--18836},
  year={2025}
}

@inproceedings{radford2021learning,
  title={Learning transferable visual models from natural language supervision},
  author={Radford, Alec and Kim, Jong Wook and Hallacy, Chris and Ramesh, Aditya and Goh, Gabriel and Agarwal, Sandhini and Sastry, Girish and Askell, Amanda and Mishkin, Pamela and Clark, Jack and others},
  booktitle={International conference on machine learning},
  pages={8748--8763},
  year={2021},
  organization={PmLR}
}

@ARTICLE{1284395,
  author={Zhou Wang and Bovik, A.C. and Sheikh, H.R. and Simoncelli, E.P.},
  journal={IEEE Transactions on Image Processing}, 
  title={Image quality assessment: from error visibility to structural similarity}, 
  year={2004},
  volume={13},
  number={4},
  pages={600-612},
  doi={10.1109/TIP.2003.819861}}

@inproceedings{song2024revisiting,
  title={Revisiting code similarity evaluation with abstract syntax tree edit distance},
  author={Song, Yewei and Lothritz, Cedric and Tang, Xunzhu and Bissyand{\'e}, Tegawend{\'e} and Klein, Jacques},
  booktitle={Proceedings of the 62nd Annual Meeting of the Association for Computational Linguistics (Volume 2: Short Papers)},
  pages={38--46},
  year={2024}
}

@article{tversky2002animation,
  title={Animation: can it facilitate?},
  author={Tversky, Barbara and Morrison, Julie Bauer and Betrancourt, Mireille},
  journal={International journal of human-computer studies},
  volume={57},
  number={4},
  pages={247--262},
  year={2002},
  publisher={Elsevier}
}

@article{kaiser1992influence,
  title={Influence of animation on dynamical judgments.},
  author={Kaiser, Mary K and Proffitt, Dennis R and Whelan, Susan M and Hecht, Heiko},
  journal={Journal of experimental Psychology: Human Perception and performance},
  volume={18},
  number={3},
  pages={669},
  year={1992},
  publisher={American Psychological Association}
}

@article{betrancourt2005animation,
  title={The animation and interactivity principles in multimedia learning},
  author={Betrancourt, Mireille},
  journal={The Cambridge handbook of multimedia learning},
  pages={287--296},
  year={2005},
  publisher={New York: Cambridge University Press}
}

@article{liang2026visphyworld,
  title={VisPhyWorld: Probing Physical Reasoning via Code-Driven Video Reconstruction},
  author={Liang, Jiarong and Ku, Max and Hui, Ka-Hei and Nie, Ping and Chen, Wenhu},
  journal={arXiv preprint arXiv:2602.13294},
  year={2026}
}

@article{lu2026worldcoder,
  title={WorldCoder-Bench: Benchmarking Physically Grounded 3D World Synthesis},
  author={Lu, Shuo and Xu, Yinuo and Yu, Kecheng and Jiang, Siru and Yu, Yongcan and Wang, Yubin and Yang, Haitao and Zhang, Yuxiang and Wang, Bin and He, Ran and others},
  journal={arXiv preprint arXiv:2606.01869},
  year={2026}
}

@article{xie2026physcodebench,
  title={Physcodebench: Benchmarking physics-aware symbolic simulation of 3d scenes via self-corrective multi-agent refinement},
  author={Xie, Tianyidan and Wang, Peiyu and Jiaxin, Hu and Qian, Yuyi and Wang, Yuxuan and Wang, Shenyi and Ma, Rui and Peng, Yanlun and Wang, Lanjun and Tai, Ying and others},
  journal={arXiv preprint arXiv:2604.23580},
  year={2026}
}

\appendix
\clearpage
\section*{Appendix}

The appendix is organized as follows:
\begin{itemize}
    \item \autoref{app:data}: Data Processing Details
    \item \autoref{app:css_properties}: CSS Animation Property Distribution
    \item \autoref{app:other-metrics}: Alternative Metrics
    \item \autoref{app:full-frame}: Evaluating with Full Frames
    \item \autoref{app:ir-variants}: Iterative Refinement Variants
    \item \autoref{app:sft-details}: Supervised Fine-tuning Details
    \item \autoref{app:fps}: Input FPS Ablation
    \item \autoref{app:human}: Human Annotation Design
    \item \autoref{app:prompts}: Prompts
\end{itemize}

\section{Data Processing Details}
\label{app:data}

\paragraph{Source} The dataset is derived from 25 public CodePen collections hand-picked to capture the diversity of decorative web animations. All scraped pens are published on CodePen under the platform's default MIT license. All source links are included in the metadata of each example in the dataset released.

\paragraph{Pipeline} Each example is collected, cleaned, and converted through the following seven steps: \\
(1)~\textit{Scrape URLs}: we collect example URLs using a script paginating through the public CodePen collections above. \\
(2)~\textit{Download}: we download each example's raw HTML, CSS, and JavaScript source files along with metadata, such as title and source link. \\
(3)~\textit{Normalize}: source files are normalized to plain HTML, CSS, and JavaScript using dedicated compilers (Pug, SASS/SCSS, and Babel). For a small subset of examples written in HAML or LESS, available compilers did not yield valid results, so we used GPT-4.1~\cite{achiam2023gpt} to translate the source into plain HTML and CSS. This is the only stage where an LLM contributes to the dataset content, and the affected examples are flagged in the Croissant metadata with \texttt{rai:hasSyntheticData = true}. \\
(4)~\textit{Extract external resources}: references to external scripts and stylesheets are extracted from the source and recorded in the \texttt{external\_resources} field of each example's \texttt{metadata.json}. \\
(5)~\textit{Render}: each example is rendered to MP4 using a headless Chromium recorder driven by Puppeteer-capture\footnote{\url{https://github.com/alexey-pelykh/puppeteer-capture}} with the protocol described below. \\
(6)~\textit{Filter}: we manually reviewed all recorded animations and dropped examples that fail to render or render to a single static frame. \\
(7)~\textit{Split}: the remaining 1,069 examples are partitioned into 769 train, 86 validation, and 214 test. Of the 1,069 examples, 172 were created by splitting 32 parent animations into smaller atomic examples, as each parent contains multiple disjoint animated elements. We make sure all animations from the same parent are grouped in the same split to prevent leakage.

\paragraph{Recording protocol} All animations are deterministically recorded by a browser recorder using headless Chromium driven by Puppeteer-capture. We use a viewport of $1024 \times 768$ pixels, recorded at 30 frames per second, with a recording window of 10~seconds. The final recordings range from 2 to 8~seconds depending on the duration of each example's animation cycle. All examples are encoded as H.264 MP4 videos with ffmpeg.

\paragraph{Dataset structure} Each example directory contains three files: \texttt{animation.mp4} (the rendered video); \texttt{full.html} (the HTML document with inline \texttt{<style>} and \texttt{<script>} tags); and \texttt{metadata.json} with three fields: \texttt{title} (the example title given by the original author), \texttt{url} (CodePen URL), and \texttt{external\_resources} (URLs of external scripts and stylesheets the example depends on).

\section{CSS Animation Property Distribution}
\label{app:css_properties}

\autoref{tab:anim_props} shows the standard CSS animation and transform properties and their distribution across the 1,069-example dataset.
The \texttt{animation} shorthand dominates at 81.6\% of examples, while \texttt{transform}appears in 72.7\%.
Sub-properties, such as \texttt{animation-delay}, \texttt{animation-duration}, \texttt{animation-timing-function}, appear in only 8--23\% of examples, indicating a strong preference for the shorthand form.
\texttt{transform-origin} and \texttt{transform-style} co-occur with \texttt{transform} in 27.4\% and 16.6\% of examples, respectively, commonly used for 3D transforms in the animations. 
\autoref{tab:impl_diversity} shows a breakdown of the implementation methods used by the examples.

\begin{table}[h]
\centering
\caption{Standard CSS animation and transform properties, sorted by prevalence across 1,069 examples.}
\label{tab:anim_props}
\begin{tabular}{lrr}
\toprule
\textbf{Property} & \textbf{examples} & \textbf{\% of Examples} \\
\midrule
\texttt{animation}                & 872 & 81.6 \\
\texttt{transform}                & 776 & 72.7 \\
\texttt{transform-origin}         & 293 & 27.4 \\
\texttt{animation-delay}          & 247 & 23.1 \\
\texttt{transform-style}          & 177 & 16.6 \\
\texttt{animation-name}           &  94 &  8.8 \\
\texttt{animation-duration}       &  92 &  8.6 \\
\texttt{animation-timing-function}&  88 &  8.2 \\
\texttt{transition}               &  77 &  7.2 \\
\texttt{animation-iteration-count}&  66 &  6.2 \\
\texttt{animation-direction}      &  42 &  3.9 \\
\texttt{animation-play-state}     &  12 &  1.1 \\
\texttt{animation-fill-mode}      &  10 &  0.9 \\
\bottomrule
\end{tabular}
\end{table}

\begin{table}[h]
\centering
\small
\caption{Implementation diversity of the reference code across 1,069 examples. Rows are not mutually exclusive. An example may contain multiple implementation methods or libraries.}
\label{tab:impl_diversity}
\begin{tabular}{lrr}
\toprule
\textbf{Implementation} & \textbf{Examples} & \textbf{\% of Examples} \\
\midrule
CSS \texttt{@keyframes}                                      & 973 & 91.0 \\
CSS \texttt{transform}                                       & 861 & 80.5 \\
Inline JavaScript                                            & 150 & 14.0 \\
Inline SVG                                                   & 122 & 11.4 \\
CSS \texttt{transition}                                      &  77 &  7.2 \\
External animation library                                   &  64 &  6.0 \\
\quad GSAP / TweenMax (MorphSVG, DrawSVG, Physics2D, MotionPath) & 45 & 4.2 \\
\quad Velocity.js                                            &  13 &  1.2 \\
\quad Three.js / WebGL                                       &   5 &  0.5 \\
\quad Lottie / Bodymovin                                     &   3 &  0.3 \\
\quad p5.js                                                  &   1 &  0.1 \\
\texttt{<canvas>} (2-D or WebGL)                             &  15 &  1.4 \\
SMIL \texttt{<animate>} elements                             &  11 &  1.0 \\
\bottomrule
\end{tabular}
\end{table}

\section{Alternative Metrics} \label{app:other-metrics}

To select automated metrics that best reflect human preferences, we ran a logistic regression study predicting pairwise human preference (overall quality, appearance, and motion) from 8 candidate metrics. Each metric is used as a feature by computing the score delta between the two compared outputs. We use the feature significance as the selection criterion.

\begin{table}[h]
\centering
\small
\caption{Logistic regression coefficients predicting pairwise human preference from metric deltas.
Significant predictors ($p<0.05$) are in \textbf{bold}. Results are from a single 50/50 split by animation (seed 42), yielding $n_\text{test}=308$ non-tied comparisons for all dimensions.}
\label{tab:metric_selection}
\setlength{\tabcolsep}{4pt}
\begin{tabular}{lp{3.4cm}cccccc}
\toprule
 & & \multicolumn{2}{c}{\textbf{Overall}} & \multicolumn{2}{c}{\textbf{Appearance}} & \multicolumn{2}{c}{\textbf{Motion}} \\
\cmidrule(lr){3-4}\cmidrule(lr){5-6}\cmidrule(lr){7-8}
\textbf{Metric} & \textbf{Description} & Coef. & $p$ & Coef. & $p$ & Coef. & $p$ \\
\midrule
CLIP\cite{radford2021learning}            & Semantic frame similarity via CLIP embeddings (DTW-aligned) & 19.90 & 0.316 & 14.96 & 0.383 & 2.81 & 0.850 \\
CLIP-DTW        & DTW cost using CLIP scores & 0.043 & 0.652 & 0.037 & 0.650 & 0.009 & 0.904 \\
DreamSim \cite{fu2023dreamsim}       & Semantic frame similarity via DreamSim embeddings (DTW-aligned) & \textbf{13.62} & \textbf{0.046} & \textbf{13.48} & \textbf{0.019} & 9.89 & 0.062 \\
DreamSim-DTW    & DTW cost using DreamSim scores & 0.008 & 0.800 & $-$0.002 & 0.955 & 0.008 & 0.737 \\
SSIM \cite{1284395}           & Pixel-level structural similarity (DTW-aligned) & 1.17 & 0.584 & $-$1.20 & 0.510 & 0.97 & 0.602 \\
SSIM-DTW        & DTW cost using SSIM scores & 0.001 & 0.871 & $-$0.003 & 0.673 & 0.002 & 0.754 \\
AST-TSED \cite{song2024revisiting}            & Code similarity based on tree edit distance using AST & 1.97 & 0.252 & $-$0.78 & 0.593 & $-$0.36 & 0.816 \\
Temporal Fidelity & Trajectory-based similarity using Chamfer distance  & \textbf{7.37} & \textbf{0.001} & \textbf{4.49} & \textbf{0.018} & \textbf{13.92} & \textbf{$<$0.001} \\
\midrule
\multicolumn{2}{l}{Test accuracy} & \multicolumn{2}{c}{0.841} & \multicolumn{2}{c}{0.825} & \multicolumn{2}{c}{0.792} \\
\bottomrule
\end{tabular}
\end{table}

\paragraph{Candidate metrics}
Table~\ref{tab:metric_selection} lists all 8 metrics with brief descriptions and their logistic regression coefficients (and $p$-values) for each human judgment dimension.

\paragraph{Analysis}
Only \textbf{DreamSim} and \textbf{temporal fidelity} are statistically significant predictors of human preference. DreamSim is significant for overall ($p=0.046$) and appearance ($p=0.019$) judgments, while temporal fidelity is strongly significant across all three dimensions (overall $p=0.001$; appearance $p=0.018$; motion $p<0.001$). 
All other metrics (CLIP, SSIM, TSED, and their DTW-aligned variants) fail to reach significance in any dimension. 
Based on these results, we adopt DreamSim as our \emph{appearance similarity} metric and temporal fidelity as our \emph{temporal similarity} metric for all main evaluations.

\section{Evaluating with Full Frames}\label{app:full-frame}

Appearance and temporal similarity, the two metrics we propose, can be computed either over the whole rendered frame or within the bounding box of the animated element. In this section, we compare both settings across five baselines at 2 FPS to justify the bounding box used in our main experiments.

\paragraph{Results}
Table~\ref{tab:fullframe_vs_animated} reports the mean and per-sample standard deviation of each metric under both settings.

\begin{table}[h]
\centering
\small
\caption{Effect of evaluation region on appearance similarity (DreamSim) and temporal similarity.
Per-sample standard deviation in parentheses. ``Animated only'' crops to the bounding box of the
animated element; ``full frame'' uses the entire rendered canvas.}
\label{tab:fullframe_vs_animated}
\setlength{\tabcolsep}{4pt}
\begin{tabular}{lcccc}
\toprule
 & \multicolumn{2}{c}{\textbf{Animated Only}} & \multicolumn{2}{c}{\textbf{Full Frame}} \\
\cmidrule(lr){2-3}\cmidrule(lr){4-5}
\textbf{Baseline} & \textbf{App.\,$\uparrow$} & \textbf{Temp.\,$\uparrow$} & \textbf{App.\,$\uparrow$} & \textbf{Temp.\,$\uparrow$} \\
\midrule
Claude (2 FPS, img)  & 0.816\,{\small(0.106)} & 0.287\,{\small(0.094)} & 0.672\,{\small(0.281)} & 0.269\,{\small(0.111)} \\
GPT (2 FPS, img)  & 0.837\,{\small(0.100)} & 0.289\,{\small(0.084)} & 0.705\,{\small(0.271)} & 0.268\,{\small(0.097)} \\
Gemini (2 FPS, vid)  & 0.799\,{\small(0.104)} & 0.295\,{\small(0.096)} & 0.633\,{\small(0.281)} & 0.271\,{\small(0.105)} \\
LLaMA (2 FPS, img)   & 0.617\,{\small(0.157)} & 0.208\,{\small(0.109)} & 0.374\,{\small(0.354)} & 0.181\,{\small(0.118)} \\
Qwen (2 FPS, vid)    & 0.673\,{\small(0.155)} & 0.234\,{\small(0.085)} & 0.487\,{\small(0.347)} & 0.188\,{\small(0.101)} \\
\bottomrule
\end{tabular}
\end{table}

\paragraph{Analysis}
Full-frame evaluation results in lower appearance scores and greater variation across examples for every baseline. The standard deviation is 2.2--2.7 times larger than with animated-only evaluation, and the mean score drops by 0.13--0.24 for every baseline. 
Differences in layout and background may contribute to this increase in variation, because the animated element often covers only a small part of the canvas.
In contrast, temporal similarity changes less, since the metric tracks only pixels that change over time (Section~\ref{sec:eval}). 
Thus, we use animated-only evaluation in all main experiments to focus on the animated element only.

\section{Iterative Refinement Variants} \label{app:ir-variants}
\begin{figure}[h]
\centering
\includegraphics[width=\linewidth]{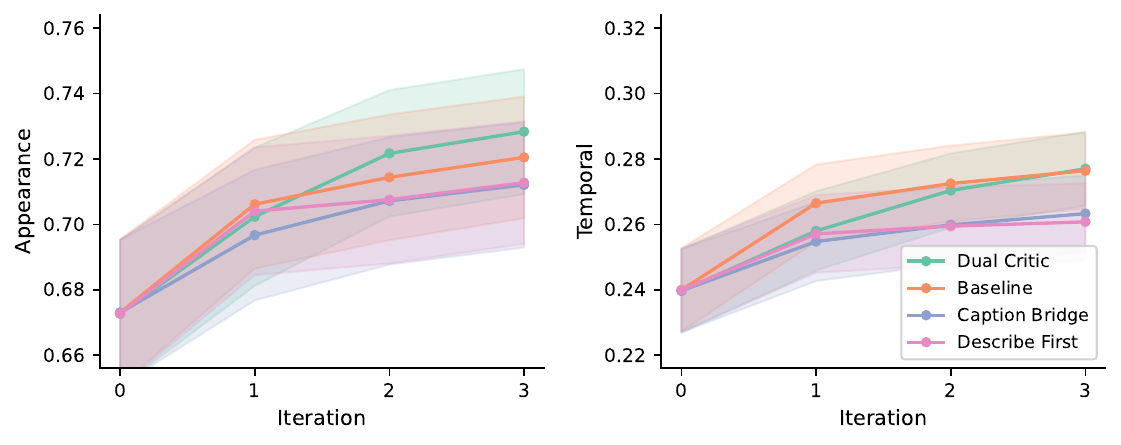}
\caption{Mean appearance and temporal similarity of the four refinement variants over three iterations, where each example contributes its best score up to that iteration. Shaded regions show 95\% confidence intervals. Dual critic outperforms the other variants on both metrics. All variants improve most in the first iteration before leveling off.}
\label{fig:ir-results}
\end{figure}

We describe and compare four iterative refinement variants, specifically a single-critic baseline (adapted from Self-Refine~\citep{madaan2023self}), describe-first (adapted from ChartIR~\citep{xu2025improved}), caption bridge (adapted from PhyT2V~\citep{xue2025phyt2v}), and dual critic (adapted from METAL~\cite{li2025metal}). We use dual critic in Section~\ref{sec:benchmarking} to measure how much iterative refinement improves over zero-shot generation, as it achieves the highest appearance and temporal similarity of the four (Figure~\ref{fig:ir-results}).

All four variants share the same basic loop: a generator $G$ writes an initial program, which is then refined over $T$ iterations. At each iteration $i \in \{1, \dots, T\}$, a critic $C$ compares the rendering of the previous program with the target animation and provides feedback on the differences. An editor $E$ then uses this feedback, together with the previous program and its rendering, to write an updated program. We use the same Qwen3-VL-8B-Instruct model for all three roles, run $T{=}3$ iterations, and preprocess videos as described in Section~\ref{sec:benchmarking}.

\paragraph{Shared initial generation} If each variant generated its own initial program, differences in performance would reflect both the refinement method and randomness in the initial generation, since $G$ samples at a nonzero temperature. To control for this, we generate one initial program per example in a single zero-shot pass and use it as the starting point for all four variants. Differences between variants therefore mainly reflect how each produces feedback.

\paragraph{Inference settings} For iterative refinement, the generator, critic, and editor decode with temperature 0.2, top-$p$ 0.9, and a maximum output length of 2048 tokens. The visual critic in the dual-critic variant instead uses temperature 0 for deterministic decoding. At each iteration, we render the program with the same headless Chromium recorder described in Section~\ref{sec:benchmarking}, so the critic sees the program as it appears in a browser.

\paragraph{Baseline} The baseline is the simplest of the four variants, and the other three each extend it with one additional component. It applies the self-refinement loop of \citet{madaan2023self} to our task, with one critique and one edit per iteration. The critic compares the target video with the rendering of the previous program and writes a free-form list of differences, which the editor uses to revise the program.

\paragraph{Describe-first} The describe-first variant follows the description-then-code approach of ChartIR~\citep{xu2025improved}. At iteration $0$, $G$ writes a structured description of the target animation, covering its objects, motion, timing, order of events, and visual effects, and then generates the initial program from this description. At each subsequent iteration, the critic receives the target video, the rendering of the previous program, and the same description, and provides feedback on the differences between the two animations. The same description is used at every refinement step.

\paragraph{Caption bridge} The caption bridge variant follows the caption-based feedback approach of PhyT2V~\citep{xue2025phyt2v}. At each iteration, $G$ writes separate captions for the target video and the rendering of the previous program. The critic compares these captions without access to the videos themselves and writes feedback on their differences. The editor then uses this feedback, together with the previous program and its rendering, to revise the program.

\paragraph{Compute resources} We ran all iterative refinement experiments on 8 NVIDIA H200 GPUs with 144GB of memory each, and we evaluated the four variants in parallel across GPUs, which took about 3 hours in total.

\section{Supervised Fine-tuning Details} \label{app:sft-details}
We fine-tune Qwen3-VL-8B-Instruct on the training split of our dataset. 
We compare two approaches: LoRA~\citep{hu2022lora} and full fine-tuning of the language component. The vision encoder remains frozen in both settings.

\paragraph{Data and preprocessing} We use the user prompt from Section~\ref{sec:benchmarking} as input and the original source code as the target. Videos are decoded with the \texttt{decord} backend and sampled at 2\,FPS, with at most $336^2$ pixels per frame ($\texttt{max\_pixels}=112896$). The combined video-and-text sequence is truncated to 16384 tokens.

\paragraph{LoRA} We attach adapters to the attention projections (\texttt{q\_proj}, \texttt{k\_proj}, \texttt{v\_proj}, \texttt{o\_proj}) and the feedforward projections (\texttt{gate\_proj}, \texttt{up\_proj}, \texttt{down\_proj}) in every transformer block of the language model, and we use rank $r{=}64$, scaling $\alpha{=}128$, and dropout $0.05$.

\paragraph{Full fine-tuning} For full fine-tuning, we update all parameters of the language model and keep the vision encoder frozen. Table~\ref{tab:sft-hparams} lists the training settings for both configurations. For LoRA we use the checkpoint with the lowest validation loss (epoch 2), and for full fine-tuning we use the final checkpoint (epoch 3), which also has the lowest validation loss.

\begin{table}[h]
\centering
\small
\caption{Training settings for the two fine-tuning configurations.}
\label{tab:sft-hparams}
\begin{tabular}{lcc}
\toprule
 & \textbf{LoRA} & \textbf{Full fine-tuning} \\
\midrule
Trained parameters & LoRA adapters & All language-model parameters \\
Optimizer & AdamW & AdamW \\
Peak learning rate & $10^{-4}$ & $10^{-5}$ \\
Schedule & Cosine, 5\% linear warmup & Cosine, 3\% warmup \\
Weight decay & 0.01 & 0.01 \\
Effective batch size & 8 (1 per GPU $\times$ 8 GPUs) & 16 (1 per GPU $\times$ 8 GPUs $\times$ 2 accum.) \\
Epochs & 5 & 3 \\
Checkpoint used & Epoch 2 (lowest val.\ loss) & Epoch 3 (final, lowest val.\ loss) \\
DeepSpeed ZeRO-3 & -- & \checkmark \\
\bottomrule
\end{tabular}
\end{table}

\paragraph{Memory} Both configurations use gradient checkpointing and the Liger fused linear-cross-entropy kernel~\citep{hsu2024liger} to reduce training memory. The Liger kernel combines the final linear projection and the cross-entropy loss into a single Triton kernel and processes the sequence in tiles, so the full logits tensor, which would not fit in GPU memory at our sequence length, is never stored. Full fine-tuning also uses DeepSpeed ZeRO-3.

\paragraph{Inference} To evaluate the fine-tuned models, we use the same prompts and video preprocessing as in training, sampling input videos at 2\,FPS with at most 112,896 pixels per frame. Both configurations use greedy decoding with a repetition penalty of 1.05 and a maximum of 16384 generated tokens. We render and record the generated programs using the same protocol as the zero-shot baselines in Section~\ref{sec:benchmarking}.

\paragraph{Compute resources} All fine-tuning experiments were run on 8 NVIDIA RTX A6000 GPUs (48\,GB each) and required approximately 1.4 hours of training time.

\section{Input FPS Ablation} \label{app:fps}
\begin{figure}[h]
\centering
\includegraphics[width=\linewidth]{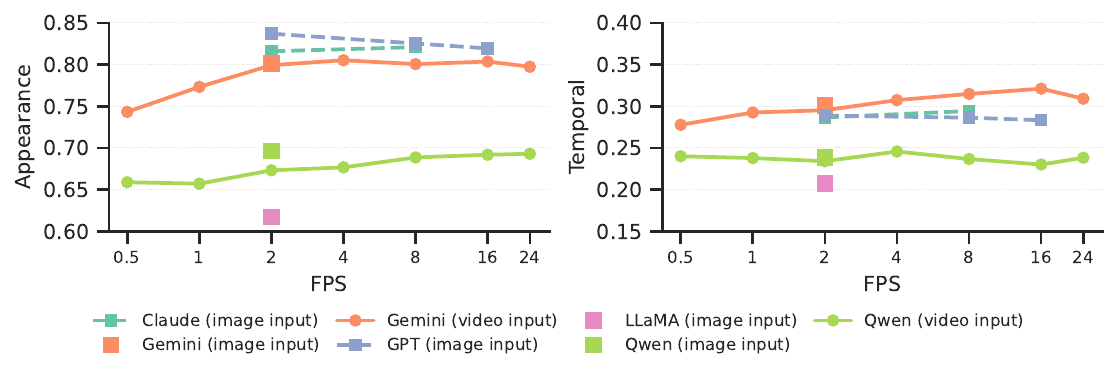}
\caption{Input FPS and modality sweep over baseline models on test set.}
\label{fig:fps}
\end{figure}

We examine how varying the number of input frames affects zero-shot baseline performance across models and input modalities. Due to API cost constraints and per-model context limits, we did not evaluate all frame rate configurations for every model. For instance, LLaMA was only evaluated at 2 FPS. Claude and GPT do not accept video input, so they were tested in image-frame mode only. For the two models that accept native video input, Gemini and Qwen, we sweep the full range from 0.5 to 24 FPS.

\paragraph{Results}
Table~\ref{tab:framerate_ablation} reports appearance similarity (DreamSim, higher is better) and temporal similarity (temporal fidelity, higher is better) across all evaluated configurations. \autoref{fig:fps} shows the trend of each model's performance over input FPS settings.

\begin{table}[h]
\centering
\small
\caption{Frame rate ablation for zero-shot baselines. $\uparrow$ = higher is better.
\textbf{Frames} is the mean number of frames the model actually receives at that rate,
averaged over the 214 test clips (mean duration 4.9\,s).}
\label{tab:framerate_ablation}
\setlength{\tabcolsep}{5pt}
\begin{tabular}{llcccc}
\toprule
\textbf{Model} & \textbf{Modality} & \textbf{FPS} & \textbf{Frames} & \textbf{Appearance Sim.\,$\uparrow$} & \textbf{Temporal Sim.\,$\uparrow$} \\
\midrule
\multirow{2}{*}{Claude} & Image & 2   & 9.4   & 0.816 & 0.287 \\
                        & Image & 8   & 38.4  & 0.821 & 0.294 \\
\midrule
\multirow{3}{*}{GPT}    & Image & 2   & 9.4   & \textbf{0.837} & 0.289 \\
                        & Image & 8   & 38.4  & 0.825 & 0.286 \\
                        & Image & 16  & 76.9  & 0.819 & 0.283 \\
\midrule
\multirow{8}{*}{Gemini} & Image & 2   & 9.4   & 0.801 & 0.301 \\
                        & Video & 0.5 & 2.6   & 0.743 & 0.278 \\
                        & Video & 1   & 4.5   & 0.773 & 0.293 \\
                        & Video & 2   & 9.4   & 0.799 & 0.295 \\
                        & Video & 4   & 19.5  & 0.805 & 0.307 \\
                        & Video & 8   & 38.4  & 0.801 & 0.315 \\
                        & Video & 16  & 76.9  & 0.804 & \textbf{0.321} \\
                        & Video & 24  & 116.3 & 0.797 & 0.309 \\
\midrule
LLaMA                   & Image & 2   & 9.4   & 0.617 & 0.208 \\
\midrule
\multirow{8}{*}{Qwen}   & Image & 2   & 9.4   & 0.696 & 0.238 \\
                        & Video & 0.5 & 2.6   & 0.659 & 0.240 \\
                        & Video & 1   & 4.5   & 0.657 & 0.238 \\
                        & Video & 2   & 9.4   & 0.673 & 0.234 \\
                        & Video & 4   & 19.5  & 0.677 & 0.246 \\
                        & Video & 8   & 38.4  & 0.689 & 0.237 \\
                        & Video & 16  & 76.9  & 0.692 & 0.230 \\
                        & Video & 24  & 116.3 & 0.693 & 0.238 \\
\bottomrule
\end{tabular}
\end{table}

\paragraph{Tradeoff between performance and input FPS}
Increasing the frame rate yields marginal improvements in both metrics for Claude but slightly decreases performance for GPT.

For Gemini, which supports both image and video input, temporal similarity increases steadily until 16 FPS, while appearance similarity plateaus early at around 4 FPS and then remains stable ($\sim$0.80). Temporal similarity increases gradually from 0.5 to 16 FPS for a total of $+0.043$ and drops at 24 FPS. The performance drops when more frames are added beyond 16 FPS possibly due to redundancy in the added frames. The animations are mostly continuous, without sudden high-frequency visual changes, so at high frame rates the input contains a large number of near-identical frames, adding context length and redundancy without new information. Gemini with image input at 2 FPS performs comparably to video input at the same frame rate, suggesting the input modality has limited impact. 

For Qwen, there is no effect of FPS on temporal similarity, which fluctuates within a narrow range (0.230--0.246) with no trend with respect to input FPS. However, its appearance score increases generally over the same range from 0.659 to 0.693. This suggests that although this model is able to reproduce input appearance more faithfully with denser input, it is not able to extract motion from these additional frames, and it remains well below frontier models regardless of input settings.

\section{Human Annotation Design} \label{app:human}

\begin{figure}[h]
\centering
\includegraphics[width=\linewidth]{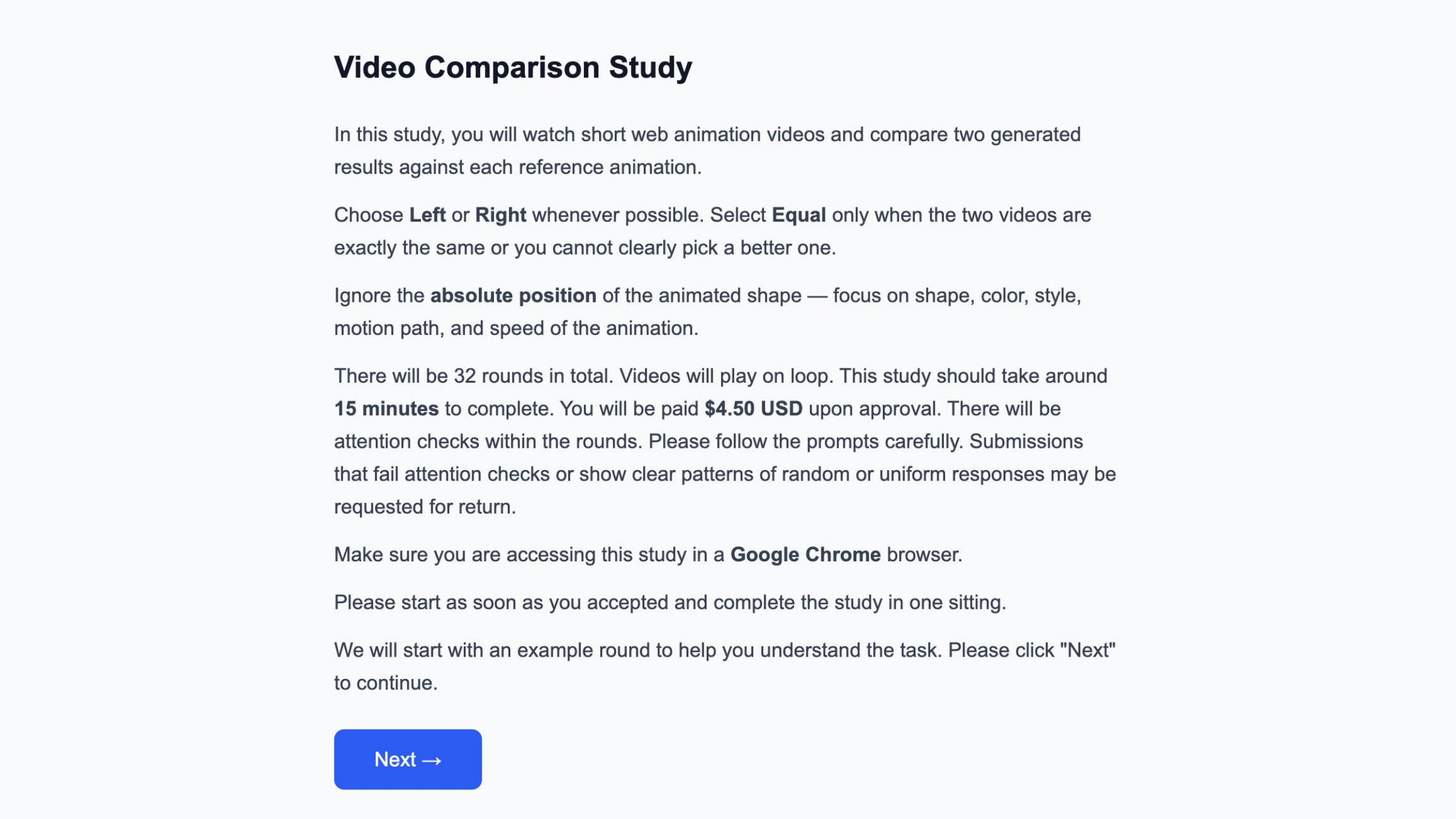}
\caption{Human annotation task interface (instructions).}
\label{fig:task-instruction}
\end{figure}

\begin{figure}[h]
\centering
\includegraphics[width=\linewidth]{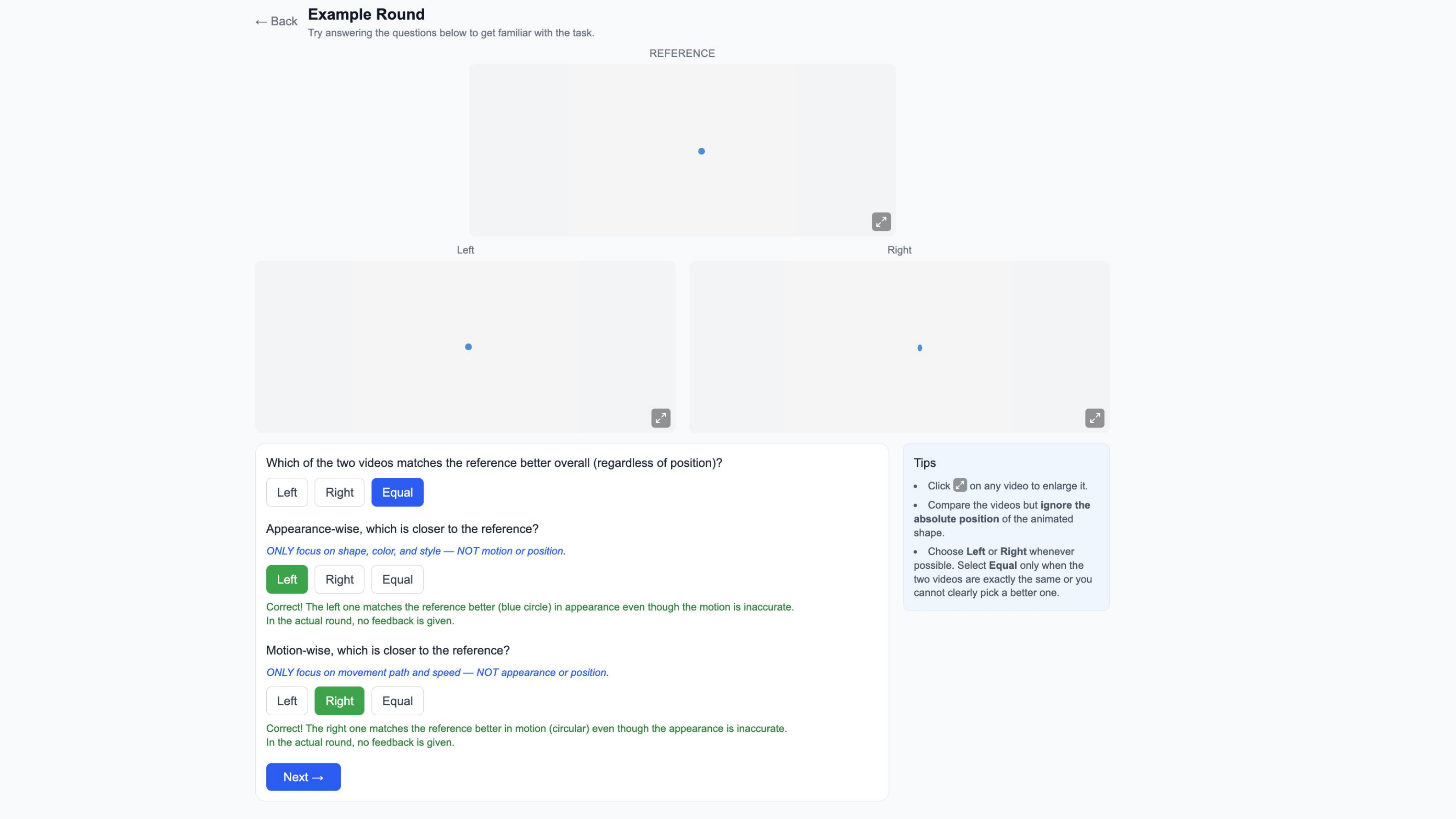}
\caption{Human annotation task interface (practice round).}
\label{fig:task-practice}
\end{figure}

\paragraph{Task design}
Annotators first read a three-page instruction. The first page (\autoref{fig:task-instruction}) describes the study. The second page presents an interactive practice round in which annotators must answer two of the three questions correctly before proceeding, with immediate correct/incorrect feedback (\autoref{fig:task-practice}). The third page restates the key rules and tips.

\begin{figure}[h]
\centering
\includegraphics[width=\linewidth]{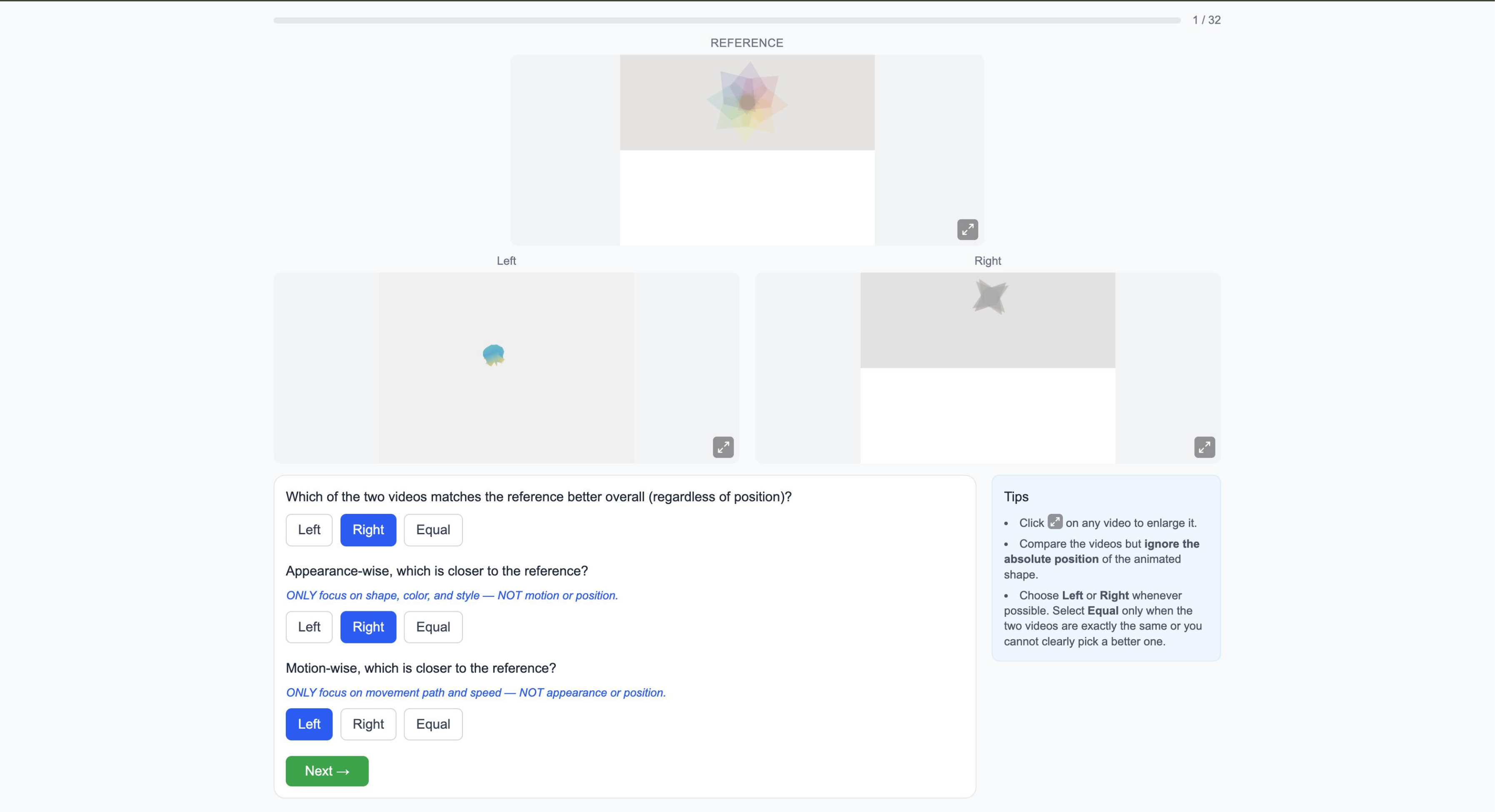}
\caption{Human annotation task interface (actual trial).}
\label{fig:task-trial}
\end{figure}

Each annotation round (\autoref{fig:task-trial}) displays a reference video above two side-by-side candidate videos labeled Left and Right. A 5-second countdown prevents answering immediately. Three questions are then shown in sequence: overall match (which candidate better matches the reference), appearance (shape, color, and style, ignoring motion), and motion (movement path and speed, ignoring appearance). If an annotator selects Equal for all three questions, a confirmation dialog prompts them to reconsider before submitting.

One \textit{implicit} attention check is inserted at a random position between rounds 1 and 30. It is visually identical to a real round, but all three videos are the same clip (the reference is used as both Left and Right), so the correct answer to every question is Equal. No indication is given that the round is a check.

One \textit{explicit} attention check is inserted after round 15. It displays a previously seen comparison, but each question shows a pre-assigned required answer in red (e.g., ``Please choose Right for this question''). The annotator must follow these instructions exactly. Whether they pass or fail the attention checks is recorded silently without immediate feedback to the annotators.

After all 32 rounds are completed, annotators are redirected to Prolific with a completion code for approval or for returning the assignment if they failed the attention checks. 

\paragraph{Data and hosting} The annotation platform is hosted on Vercel, and annotation results are stored in Firestore. Only Left/Right/Equal labels and timestamps are stored. No PII is stored.

\paragraph{Recruitment}
We recruited 93 annotators from Prolific \cite{prolific2026}. 10 failed both attention checks, so their submissions are requested for return. 18 passed the explicit attention check but failed the implicit one. These 18 annotators received partial compensation (\$2.00) for their participation. The remaining 65 annotators received full compensation for \$4.50. The task is estimated to take 15 minutes, so the annotators were paid at a rate of \$18.00 per hour.


\clearpage

\section{Prompts} \label{app:prompts}
This section lists the prompts used in all experiments. We replace the bracketed placeholders in each prompt, such as \texttt{\{current\_code\}} and \texttt{\{feedback\}}, with the corresponding inputs.
\subsection{Zero-Shot Generation}

For models that accept native video input, we pair the user message with the target animation video. For frame input, we supply a uniformly subsampled sequence of frames and use \texttt{USER\_PROMPT\_IMAGES}. This prompt is identical except for its opening clause, ``Watch this animation carefully (shown as a sequence of frames)\ldots''.

\begin{tcolorbox}[colback=gray!10!white, colframe=gray!70!black, title=System Prompt]
You are an expert web developer specializing in CSS animations and
JavaScript. Your task is to analyze an animation video and generate
the HTML/CSS/JavaScript code that recreates it. Focus on accuracy -
match the timing, easing, colors, and movement as closely as possible.
\end{tcolorbox}

\begin{tcolorbox}[colback=gray!10!white, colframe=gray!70!black, title=User Prompt]
Watch this [animation video | animation (shown as a sequence of frames)] carefully and generate the complete code
to recreate it.

Provide your response as a single, complete HTML document only.
Include all CSS inside <style> tags and any JavaScript inside <script>
tags.
Keep the code concise --- avoid generating hundreds of repetitive
CSS rules.

Be precise about:

- Animation duration and timing functions

- Colors and gradients

- Transform properties (rotate, scale, translate)

- Keyframe percentages

- Any JavaScript timing or interactions
\end{tcolorbox}

\subsection{Iterative Refinement: Shared Editor}

All four refinement variants share the same editor prompts. The editor receives the previous program and variant-specific feedback and returns a revised program.

\begin{tcolorbox}[colback=gray!10!white, colframe=gray!70!black, title=System Prompt]
You are an expert web developer. You will receive an HTML/CSS/JavaScript
animation document and specific visual feedback on how to improve it.
Your job is to produce a revised, complete HTML document that addresses
all feedback.
\end{tcolorbox}

\begin{tcolorbox}[colback=gray!10!white, colframe=gray!70!black, title=Editor User Prompt]

\{target\_video\_note\}Here is the current HTML code:

\verb|```html|

\{current\_code\}

\verb|```|

Here is the visual feedback on what needs to change:

\{feedback\}

Generate an improved version of the complete HTML document that
addresses all the feedback above. Output only a single, complete
HTML document --- no explanation, no markdown outside the document.

\end{tcolorbox}

\noindent
\texttt{\{target\_video\_note\}} is the empty string when the target video is not
re-attached on this iteration, and a short reminder sentence otherwise.

\subsection{Baseline (Single-Critic Self-Refine)}

\begin{tcolorbox}[colback=gray!10!white, colframe=gray!70!black, title=Critic System Prompt]
You are a visual expert who compares web animations. Your job is to
identify differences between a target animation and a current attempt,
then describe exactly what code changes would make them match.
\end{tcolorbox}

\begin{tcolorbox}[colback=gray!10!white, colframe=gray!70!black, title=Critic User Prompt]
You are given two animation videos:
1. TARGET -{}- the animation we want to recreate.
2. CURRENT -{}- the latest generated attempt.

Analyze them carefully and provide:
(a) A concise numbered list of specific visual differences
(colors, shapes, sizes, positions, motion paths, timing,
easing, etc.).
(b) For each difference, a concrete suggestion for what to change
in the code.

Be specific and actionable. Prioritize the most impactful fixes first.
Do NOT include any code -{}- only describe what needs to change and why.
\end{tcolorbox}

\subsection{Describe-First (ChartIR Adaptation)}

We generate the description once, at iteration~0, and pass it to the code-generation prompt to produce the initial program. The holistic-difference critic receives the same description at every subsequent iteration.

\begin{tcolorbox}[colback=gray!10!white, colframe=gray!70!black, title=Describe System Prompt]
You are an expert at analyzing web animations. Describe animations in
precise, structured detail that a developer could use to write the code.
\end{tcolorbox}

\begin{tcolorbox}[colback=gray!10!white, colframe=gray!70!black, title=Describe User Prompt]
Watch this animation video carefully and provide a structured
description:

1. OBJECTS: List every visual element (shapes, text, images) with
their colors, sizes, and initial positions.
2. MOTION: For each moving element, describe the trajectory,
direction, and distance of movement.
3. TIMING: Describe duration, delays, easing functions (linear,
ease-in, ease-out, bounce, etc.), and whether animations loop.
4. SEQUENCE: What happens first, second, third? Are things
simultaneous or sequential?
5. EFFECTS: Any gradients, shadows, opacity changes, scale changes,
rotations?

Be extremely precise about numbers (pixels, seconds, degrees).
\end{tcolorbox}

\begin{tcolorbox}[colback=gray!10!white, colframe=gray!70!black, title=Describe-Then-Code Prompt]
Here is a detailed description of an animation:

\{description\}

Generate a single, complete HTML document that recreates this
animation exactly.
Include all CSS in <style> tags and JavaScript in <script> tags.
Match the timing, easing, colors, and movement as precisely as
possible.
\end{tcolorbox}

\begin{tcolorbox}[colback=gray!10!white, colframe=gray!70!black, title=Holistic-Difference Critic Prompt]
You are given two animation videos:
1. TARGET -{}- the animation we want to recreate.
2. CURRENT -{}- the latest generated attempt.

And the original structured description of the target:
\{description\}

Provide a HOLISTIC difference description covering ALL dimensions
simultaneously:
- Objects: missing, extra, wrong color/size?
- Motion: wrong direction, speed, trajectory?
- Timing: wrong duration, easing, sequence?
- Effects: missing shadows, opacity, transforms?

For each difference, suggest a specific code change.
Do NOT include code.
\end{tcolorbox}

\subsection{Caption Bridge (PhyT2V Adaptation)}

We use the same caption prompt to describe the target and current videos separately. We then pass only the two captions to the comparison prompt, without the videos themselves.

\begin{tcolorbox}[colback=gray!10!white, colframe=gray!70!black, title=Caption System Prompt]
You are an expert at describing web animations in precise detail.
Describe exactly what you see happening in the animation.
\end{tcolorbox}

\begin{tcolorbox}[colback=gray!10!white, colframe=gray!70!black, title=Caption User Prompt]
Watch this animation video and describe what happens frame by frame:
- What objects are present and what they look like
- How they move, rotate, scale, or change
- The timing and sequence of events
- Any color changes, fading, or effects

Be precise about directions (left/right/up/down), speeds (fast/slow),
and timing (first 2 seconds, then...).
\end{tcolorbox}

\begin{tcolorbox}[colback=gray!10!white, colframe=gray!70!black, title=Caption Comparison Prompt]
Here are descriptions of two animations:

TARGET animation (what we want):
\{target\_caption\}

CURRENT animation (what we generated):
\{generated\_caption\}

Compare these descriptions and list the specific mismatches:
1. What is different about the objects?
2. What is different about the motion/movement?
3. What is different about the timing?
4. What is different about the visual effects?

For each mismatch, suggest a specific code change to fix it.
\end{tcolorbox}

\subsection{Dual Critic (METAL Adaptation)}

\begin{tcolorbox}[colback=gray!10!white, colframe=gray!70!black, title=Visual Critic System Prompt]
You are a visual expert. Compare two animation videos and describe
ONLY the visual differences you see. Focus on what looks different,
not code.
\end{tcolorbox}

\begin{tcolorbox}[colback=gray!10!white, colframe=gray!70!black, title=Visual Critic User Prompt]
Compare these two animations:
1. TARGET -{}- what we want.
2. CURRENT -{}- what we have.

List ONLY visual differences: colors, shapes, sizes, positions,
motion paths, speed, timing. Be specific (e.g., 'circle is blue
instead of red', 'rotation is clockwise instead of counter-clockwise').
\end{tcolorbox}

\begin{tcolorbox}[colback=gray!10!white, colframe=gray!70!black, title=Code Critic System Prompt]
You are a code review expert for CSS animations and JavaScript.
Analyze code structure and suggest improvements.
\end{tcolorbox}

\begin{tcolorbox}[colback=gray!10!white, colframe=gray!70!black, title=Code Critic User Prompt]
Here is HTML/CSS/JS code for an animation that doesn't quite match
the target.

Visual differences identified:
\{visual\_feedback\}

Current code:

\verb|```html|

\{current\_code\}

\verb|```|

For each visual difference, identify the specific CSS property,
keyframe, or JS line that causes it, and suggest the exact fix.
Be precise about values (colors, durations, transforms).
\end{tcolorbox}

\begin{tcolorbox}[colback=gray!10!white, colframe=gray!70!black, title=Dual Editor User Prompt]
Here is the current HTML code:

\verb|```html|

\{current\_code\}

\verb|```|

Visual feedback:
\{visual\_feedback\}

Code-level fixes:
\{code\_feedback\}

Generate an improved version that addresses all feedback.
Output only a single, complete HTML document.
\end{tcolorbox}

\subsection{VLM Judge for Predicting Pairwise Human Preference} \label{app:vlm-judge-prompt}

The judge receives three videos in order, the reference followed by Video A and Video B, and then the user message. 

\begin{tcolorbox}[colback=gray!10!white, colframe=gray!70!black, title=System Prompt]
You are an expert judge evaluating web animation videos.

You will see three videos: a Reference animation, Video A, and Video B.

Your task is to compare Video A and Video B against the Reference.

Ignore the absolute position of animated shapes --- focus on shape, color, style, motion path, and speed.
\end{tcolorbox}

\begin{tcolorbox}[colback=gray!10!white, colframe=gray!70!black, title=User Prompt]
Watch the three videos above: Reference (first), Video A (second), Video B (third).

Answer these three questions. For each, choose "A", "B", or "TIE".

Use TIE only when the two are indistinguishable or you genuinely cannot pick one.

Q1 (overall): Which of Video A or Video B matches the Reference better overall?

Q2 (appearance): Appearance-wise, which is closer to the Reference? Focus ONLY on shape, color, and style --- ignore motion and position.

Q3 (motion): Motion-wise, which is closer to the Reference? Focus ONLY on movement path and speed --- ignore appearance and position.

Respond ONLY with valid JSON, no other text:

\{

\hspace*{1em}"overall":    \{"winner": "A or B or TIE", "reason": "1-2 sentences"\},

\hspace*{1em}"appearance": \{"winner": "A or B or TIE", "reason": "1-2 sentences"\},

\hspace*{1em}"motion":     \{"winner": "A or B or TIE", "reason": "1-2 sentences"\}

\}
\end{tcolorbox}

\subsection{Adapting the WebVR-IM Metric} \label{app:webvr-im}

We reproduced WebVR's Interaction-and-Motion (IM) metric~\cite{dai2026webvr} from released code. We keep all prompts and code unchanged, except for two things: (1) WebVR synthesizes rubrics from the caption that generated their synthetic pages, while for our data, we caption the reference video and code instead; and (2) we judge with Gemini 3.1 Pro, to compare with our existing VLM-as-a-judge baseline, rather than Kimi-K2.5.

\paragraph{Adapted Prompt} This prompt is used to generate the caption for change (1). It receives the reference video and the user message below.

\begin{tcolorbox}[colback=gray!10!white, colframe=gray!70!black, title=Caption System Prompt]
You are a senior web designer writing the design brief that a front-end developer will implement.

You are given a video of a rendered web animation and the HTML/CSS/JS that produced it.

Write a detailed prose description of this page as a design brief, covering:

- Overall visual style: background, color palette, shapes, typography, mood.

- Every visible element: what it is, where it sits, its size, color, and shape.

- The motion: exactly what moves, along what path, in what direction, how fast, in what
order, whether it loops, and how it eases. Describe timing and sequencing concretely.

Write it as flowing prose describing the finished page, in the present tense, as if
specifying a page to be built. Do NOT mention CSS class names, HTML tags, code, keyframe
names, or the fact that you were shown a video. Do not use bullet points or headings.
\end{tcolorbox}

\begin{tcolorbox}[colback=gray!10!white, colframe=gray!70!black, title=Caption User Prompt]
HTML/CSS/JS that produced this animation:

\{html\}
\end{tcolorbox}

\paragraph{WebVR Prompts} The remaining prompts are from WebVR. The rubric is synthesized from the caption, then pruned by judging the reference video against it, dropping the criteria the reference does not meet. Each generated video and its code are then judged against the pruned rubric.

\begin{tcolorbox}[colback=gray!10!white, colframe=gray!70!black, breakable, title=Rubric Synthesis System Prompt]
You are an expert UI/UX Design Auditor and Pixel-Perfect Visual QA Specialist. Your goal is to analyze a design description and generate a **High-Granularity Structured Visual Verification Rubric**.

\#\#\# Context

You will receive a detailed description of a website. Your output will be used to programmatically validate a rendered webpage. The validation is purely visual (what can be seen), not code-based.

\#\#\# Task

Convert the description into a nested JSON tree of binary (Yes/No) criteria, categorized strictly by the schema defined below.

\#\#\# Critical Guidelines (Strict Enforcement)

1. **Visual Proof Only**: Ask "Is X visible?", "Is Y aligned left?", "Is the background black?". DO NOT ask about CSS classes, HTML tags, or internal logic.

2. **Extreme Atomicity (Rule of One)**:

\hspace*{1.5em}* **One Attribute Per Check**: Never bundle multiple attributes into one question.

\hspace*{2.5em}* *Bad:* "Is the button a red rounded rectangle with white text?"

\hspace*{2.5em}* *Good:* (Split into 3 checks):

\hspace*{3.5em}1. "Is a button visible?"

\hspace*{3.5em}2. "Is the button background red?"

\hspace*{3.5em}3. "Does the button have rounded corners?"

\hspace*{3.5em}4. "Is the button text color white?"

\hspace*{1.5em}* **No Conjunctions**: Avoid using "and", "or", "with" in your criteria strings.

3. **Decomposition Strategy**: For every major element mentioned, generate separate checks for:

\hspace*{1.5em}* **Existence/Content**: Is it there? Does it say the right text?

\hspace*{1.5em}* **Layout/Position**: Is it centered? Is it fixed to the top? Is it next to X?

\hspace*{1.5em}* **Style/Appearance**: Colors, shapes, fonts (serif/sans), contrast, opacity.

4. **Fixed Taxonomy**: You must sort all criteria into the 4 fixed top-level categories defined in the Output Format.

\#\#\# Output Format Rules

1. Output **ONLY** a valid JSON object.

2. **WRAP** the JSON content strictly between \verb|`<rubric>`| and \verb|`</rubric>`| tags.

3. **DO NOT** use Markdown code blocks (e.g., \verb|```json|). Just raw text inside the tags.

4. **SCHEMA**: You must follow this exact JSON structure:

<rubric>

\{

\hspace*{1em}"Global\_Aesthetics": [

\hspace*{2em}// List specific criteria about color palette, typography styles, and overall mood.

\hspace*{1em}],

\hspace*{1em}"Navigation\_and\_Footer": [

\hspace*{2em}// List criteria regarding the header, menu, and footer elements.

\hspace*{1em}],

\hspace*{1em}"Section\_Specific\_Layouts": \{

\hspace*{2em}// Create a key for each specific section mentioned in the text (e.g., "Hero\_Section", "About\_Section").

\hspace*{2em}// The value is a list of criteria specific to that section's layout and content.

\hspace*{2em}"Name\_of\_Section\_1": [ ... ],

\hspace*{2em}"Name\_of\_Section\_2": [ ... ]

\hspace*{1em}\},

\hspace*{1em}"Interaction\_and\_Motion": [

\hspace*{2em}// List criteria regarding animations, hover states, cursor behaviors, and transitions.

\hspace*{1em}]

\}

</rubric>

\#\#\# Few-Shot Example (Observe the Atomicity)

**Input Description:**

> "A minimalist art gallery page. The background is a warm beige 'sand' color. Navigation is hidden behind a 'hamburger' menu icon in the top right. The main title 'ECHOES' is centered. The Hero section shows a single large painting. Below that, a 'Gallery Grid' section displays square images. When scrolling, images fade in from the bottom."

**Output:**

<rubric>

\{

\hspace*{1em}"Global\_Aesthetics": [

\hspace*{2em}"Is the overall page background color a warm beige or sand tone?",

\hspace*{2em}"Is the background color solid (not a gradient or image)?",

\hspace*{2em}"Is the design style minimalist with generous whitespace?",

\hspace*{2em}"Is there high contrast between text and background?"

\hspace*{1em}],

\hspace*{1em}"Navigation\_and\_Footer": [

\hspace*{2em}"Is the standard navigation menu hidden by default?",

\hspace*{2em}"Is a menu icon visible?",

\hspace*{2em}"Does the menu icon consist of three stacked lines (hamburger style)?",

\hspace*{2em}"Is the menu icon positioned in the top-right corner?"

\hspace*{1em}],

\hspace*{1em}"Section\_Specific\_Layouts": \{

\hspace*{2em}"Hero\_Section": [

\hspace*{3em}"Is the text 'ECHOES' visible?",

\hspace*{3em}"Is the title 'ECHOES' horizontally centered?",

\hspace*{3em}"Is the title text uppercase?",

\hspace*{3em}"Is a painting image visible in this section?",

\hspace*{3em}"Is the painting displayed as a single large element?"

\hspace*{2em}],

\hspace*{2em}"Gallery\_Grid\_Section": [

\hspace*{3em}"Is there a grid layout containing multiple images?",

\hspace*{3em}"Are the images arranged in columns and rows?",

\hspace*{3em}"Do the images appear to have a square aspect ratio (1:1)?"

\hspace*{2em}]

\hspace*{1em}\},

\hspace*{1em}"Interaction\_and\_Motion": [

\hspace*{2em}"Do images initially appear with an opacity change (fade-in)?",

\hspace*{2em}"Does the animation translate from bottom to top (move up)?",

\hspace*{2em}"Is the animation triggered by the scroll action?"

\hspace*{1em}]

\}

</rubric>
\end{tcolorbox}

\begin{tcolorbox}[colback=gray!10!white, colframe=gray!70!black, title=Rubric Synthesis User Prompt]
**Input Description:**

> "\{caption\}"
\end{tcolorbox}

\begin{tcolorbox}[colback=gray!10!white, colframe=gray!70!black, title=Rubric Pruning System Prompt]
You are an expert Visual QA Engineer and Automated Testing System. Your task is to evaluate a video recording of a rendered webpage against a strict visual rubric.

\#\#\# Context

You will be provided with:

1. A **Video** showing a rendered webpage (which may include scrolling or interactions).

2. A **Rubric** (in JSON format) containing strictly categorized, atomic Yes/No questions about the visual appearance, layout, and behavior of the page.

\#\#\# Task

Watch the video carefully and answer every single question in the rubric based EXCLUSIVELY on what you can visually see in the video.

\#\#\# Critical Guidelines

1. **Exhaustive Evaluation**: You MUST evaluate every single criterion present in the input rubric. Do not skip or omit any questions.

2. **Strictly Visual Proof**: Base your answers solely on the visual evidence in the video. If a criterion asks about an animation or interaction, watch the video to see if that behavior occurs.

3. **Binary Judgment**: For each criterion, determine if it is met (\verb|`true`|) or not met (\verb|`false`|). If something is completely invisible or ambiguous in the video, mark it as \verb|`false`|.

4. **Provide Evidence**: Provide a brief, factual \verb|`reason`| detailing exactly what you saw in the video that led to your \verb|`true/false`| judgment.

\#\#\# Output Format

1. Output ONLY a valid JSON object.

2. WRAP the JSON content strictly between \verb|`<evaluation>`| and \verb|`</evaluation>`| tags.

3. Your output JSON must mirror the exact category structure (Global\_Aesthetics, Navigation\_and\_Footer, etc.) of the input rubric.

4. Replace the original criterion strings with an object containing \verb|`criterion`|, \verb|`met`| (boolean), and \verb|`reason`| (string).
\end{tcolorbox}

\begin{tcolorbox}[colback=gray!10!white, colframe=gray!70!black, title=Rubric Pruning User Prompt]
\#\#\# Rubric JSON:

\{rubric\_json\}
\end{tcolorbox}

\begin{tcolorbox}[colback=gray!10!white, colframe=gray!70!black, title=Judge System Prompt]
You are an expert Full-Stack Visual QA Engineer and Code Auditor. Your task is to evaluate a generated UI against a strict visual rubric.

\#\#\# Context

You will be provided with:

1. A **Video** showing the visual rendering of the webpage.

2. The **Source Code** (HTML/CSS/JS) used to generate the page.

3. A **Rubric** (in JSON format) containing atomic Yes/No questions about the UI.

\#\#\# Task

Evaluate every question in the rubric. You MUST strictly follow this visual-first hierarchy:

* **Primary Truth (Visuals)**: The Video is your absolute primary source of truth. Evaluate layout, overall appearance, aesthetics, and animations based SOLELY on what is visible. What the user actually sees is all that matters.

* **Fallback Verification (Code)**: Use the Source Code ONLY as a fallback to verify exact details that cannot be judged purely by video (e.g., exact HEX color codes, specific font-family names, or unseen interactive states like \verb|`hover`| effects). Do NOT use the code to override a poor visual result.

\#\#\# Critical Guidelines

1. **Zero Tolerance for Rendering Failures**: If the video displays a nearly blank screen, severe layout breakage, overlapping unreadable elements, or clearly missing CSS styling, you MUST grade the aesthetic and layout criteria as \verb|`false`|.

2. **Visual Precedence**: If the code contains an implementation, but it is invisible, broken, or visually poor in the video, it is a FAILURE. The Video always overrides the Code.

3. **Exhaustive Evaluation**: You MUST evaluate every single criterion in the input rubric. Do not skip any.

4. **Binary Judgment**: Determine if the criterion is completely met (\verb|`true`|) or not (\verb|`false`|).

5. **Factual Reasoning**: Write a concise \verb|`reason`|. Explicitly state what you observed in the **video** first. Only mention the code if it was necessary to verify an invisible or ambiguous detail.

\#\#\# Output Format

1. Output ONLY a valid JSON object.

2. WRAP the JSON content strictly between \verb|`<evaluation>`| and \verb|`</evaluation>`| tags.

3. Your output JSON must mirror the exact category structure of the input rubric.

4. Replace the original criterion strings with an object containing \verb|`criterion`|, \verb|`met`| (boolean), and \verb|`reason`| (string).
\end{tcolorbox}

\begin{tcolorbox}[colback=gray!10!white, colframe=gray!70!black, title=Judge User Prompt]
\#\#\# Rubric JSON:

\{rubric\_json\}

\#\#\# HTML Code:

\{html\}
\end{tcolorbox}


\end{document}